\documentclass[preprint,12pt]{elsarticle}

\makeatletter
\def\ps@pprintTitle{%
  \let\@oddhead\@empty
  \let\@evenhead\@empty
  \let\@oddfoot\@empty
  \let\@evenfoot\@empty}
\makeatother

\usepackage{graphicx}%
\usepackage{multirow}%
\usepackage{amsmath,amssymb,amsfonts}%
\usepackage{amsthm}%
\usepackage{mathrsfs}%
\usepackage[title]{appendix}%
\usepackage{xcolor}%
\usepackage{textcomp}%
\usepackage{manyfoot}%
\usepackage{booktabs}%
\usepackage{algorithm}%
\usepackage{algorithmicx}%
\usepackage{algpseudocode}%
\usepackage{listings}%

\usepackage{orcidlink}
\usepackage{bm}
\usepackage{colortbl}
\usepackage{xspace}
\usepackage{hyperref}
\usepackage{subcaption}
\usepackage{rotating}
\usepackage{makecell}

\usepackage{amssymb}
\usepackage{amsmath}
\usepackage{algorithm}%
\usepackage{algorithmicx}%
\usepackage{amsthm}
\newcommand{\eg}{\textit{e.g.}\@\xspace}

\usepackage{lineno}

\begin{document}

\begin{frontmatter}

\title{Instance-Aware Test-Time Segmentation \\for Continual Domain Shifts}

\author[1]{Seunghwan Lee\fnref{equal1}}
\ead{simon2@skku.edu}

\author[1]{Inyoung Jung\fnref{equal1}}
\ead{jiy03150@skku.edu}

\author[2]{Hojoon Lee}
\ead{hojoonlee1995@gmail.com}

\author[1]{Eunil Park}
\ead{eunilpark@skku.edu}

\author[1]{\\Sungeun Hong\corref{cor1}}

\affiliation[1]{organization={Department of Immersive Media Engineering},
            addressline={Sungkyunkwan University}}

\affiliation[2]{organization={Department of Electrical and Computer Engineering},
            addressline={Inha University}}

\fntext[equal1]{These authors contributed equally to this work.}
\cortext[cor1]{Corresponding author: \texttt{csehong@skku.edu} (Sungeun Hong)}

\begin{abstract}
Continual Test-Time Adaptation (CTTA) enables pre-trained models to adapt to continuously evolving domains. Existing methods have improved robustness but typically rely on fixed or batch-level thresholds, which cannot account for varying difficulty across classes and instances. This limitation is especially problematic in semantic segmentation, where each image requires dense, multi-class predictions.  
We propose an approach that adaptively adjusts pseudo labels to reflect the confidence distribution within each image and dynamically balances learning toward classes most affected by domain shifts. This fine-grained, class- and instance-aware adaptation produces more reliable supervision and mitigates error accumulation throughout continual adaptation. Extensive experiments across eight CTTA and TTA scenarios, including synthetic-to-real and long-term shifts, show that our method consistently outperforms state-of-the-art techniques, setting a new standard for semantic segmentation under evolving conditions.
\end{abstract}


\end{frontmatter}

\section{Introduction}\label{sec1}

Test-Time Adaptation (TTA) enables pre-trained models to adapt to unseen domains during inference without access to source data, making it crucial for privacy-sensitive or resource-limited applications. In contrast, Test-Time Training (TTT)\cite{sun2020test, gandelsman2022test, sinha2023test} requires source data and often relies on self-supervised objectives. Among TTA methods, entropy minimization\cite{wang2020tent, boudiaf2022parameter, goyal2022conjugate} reduces prediction uncertainty by updating batch normalization statistics, while self-supervised adaptation~\cite{zhang2022memo, niu2022efficient, chen2022contrastive} enforces consistency across augmented views or leverages pseudo labels to preserve feature stability. However, these methods assume a static target distribution, which rarely holds in real-world scenarios where the target domain evolves over time.

\begin{figure}[]
    \centering
    
    \begin{minipage}{0.25\linewidth}
        \includegraphics[width=\linewidth]{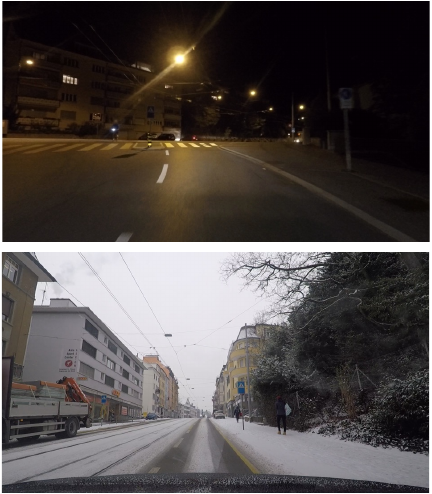}
        \centering
        (a) input image
    \end{minipage}
    \hspace{-0.1cm}
    \begin{minipage}{0.25\linewidth}
        \includegraphics[width=\linewidth]{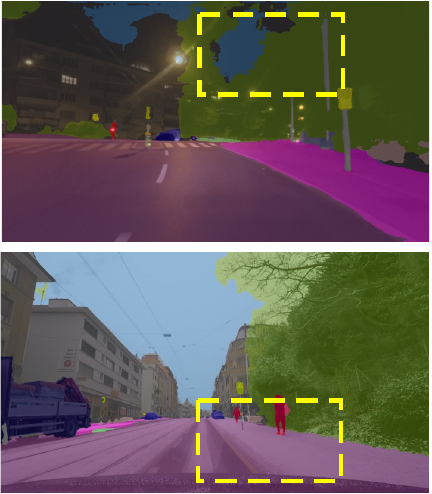}
        \centering
        (b) prediction
    \end{minipage}
    \hspace{-0.1cm}
    \begin{minipage}{0.25\linewidth}
        \includegraphics[width=\linewidth]{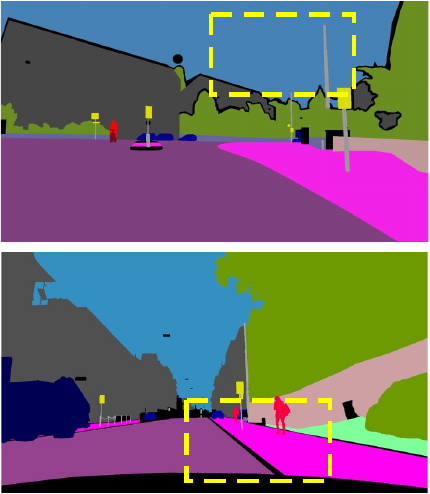}
        \centering
        (c) ground truth
    \end{minipage}
    \hspace{-0.1cm}
    \begin{minipage}{0.07\linewidth}
        \vspace{-5mm} %
        \includegraphics[width=\linewidth]{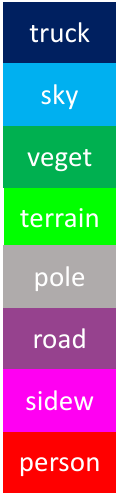}
    \end{minipage}

    \caption{ 
 Examples of semantic segmentation challenges under varying domain shifts, highlighting class-specific variations. (a) Input images captured under nighttime and snowy conditions. (b) Model predictions reveal distinct class-specific challenges: under streetlights at night, the sky is misclassified as trees, while in snowy conditions, snow-covered roads are confused with sidewalks. (c) Ground truth annotations.
    }
    \label{fig:hard_example}
    \vspace{-0.3cm}
\end{figure}

Recently, Continual Test-Time Adaptation (CTTA) addresses this issue by updating models continuously to handle non-stationary environments~\cite{wang2022continual}. CoTTA~\cite{wang2022continual}, a representative work, introduced a mean-teacher framework~\cite{tarvainen2017mean} with confidence-thresholded pseudo labeling, inspiring subsequent methods that improve robustness~\cite{dobler2023robust, yuan2023robust}, efficiency~\cite{song2023ecotta, ni2024distribution}, and flexibility~\cite{gan2023decorate, yang2024exploring}. Despite these efforts, existing CTTA methods still rely on fixed or batch-level thresholds that fail to capture how domain shifts vary across classes and instances. Dynamic thresholding techniques such as FreeMatch~\cite{wang2022freematch}, FlexMatch~\cite{zhang2021flexmatch}, CPCL~\cite{fan2023conservative}, and HRDA~\cite{hoyer2022hrda} have advanced pseudo labeling in semi-supervised and domain adaptation tasks, but they operate at the image or batch level, ignore temporal evolution, and are not designed for dense prediction. These limitations lead to misaligned pseudo labels and error accumulation under class- and instance-specific shifts. In semantic segmentation, where each image involves thousands of pixel-level predictions across diverse categories, such errors can severely degrade adaptation performance~\cite{luo2024kill, luo2024crots, wang2024search}. Figure~\ref{fig:hard_example} illustrates this: nighttime lighting causes sky regions to be confused with vegetation, and snow occlusion leads sidewalks to be misclassified as roads. These challenges call for an approach that adapts at the pixel level, considers class- and instance-specific variations, and remains robust under temporal domain changes.

\begin{figure}[]
    \centering
    \begin{subfigure}[b]{0.20\linewidth}
        \includegraphics[width=\linewidth]{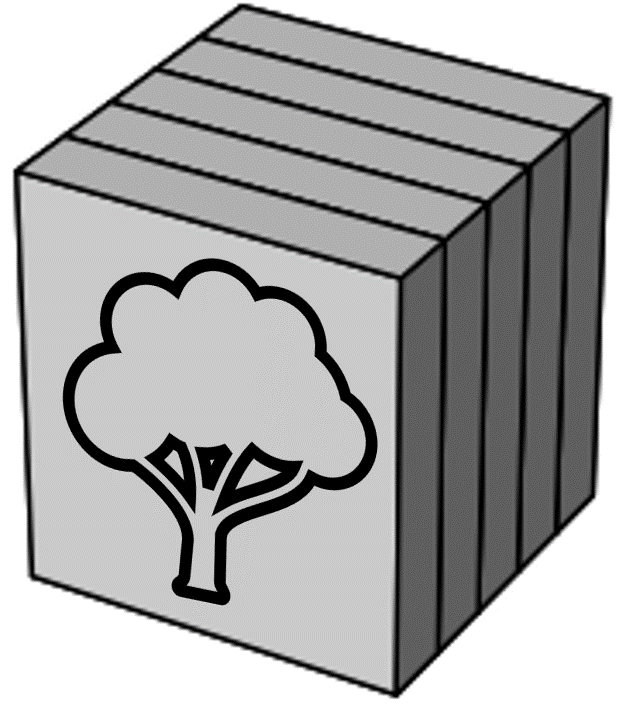}
        \subcaption{}
    \end{subfigure}
    \begin{subfigure}[b]{0.20\linewidth}
        \includegraphics[width=\linewidth]{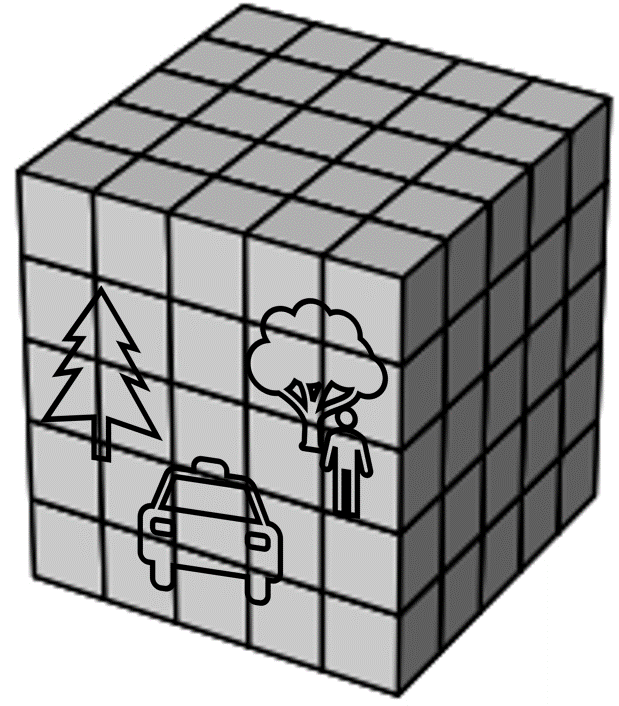}
        \subcaption{}
    \end{subfigure}
    \begin{subfigure}[b]{0.20\linewidth}
        \includegraphics[width=\linewidth]{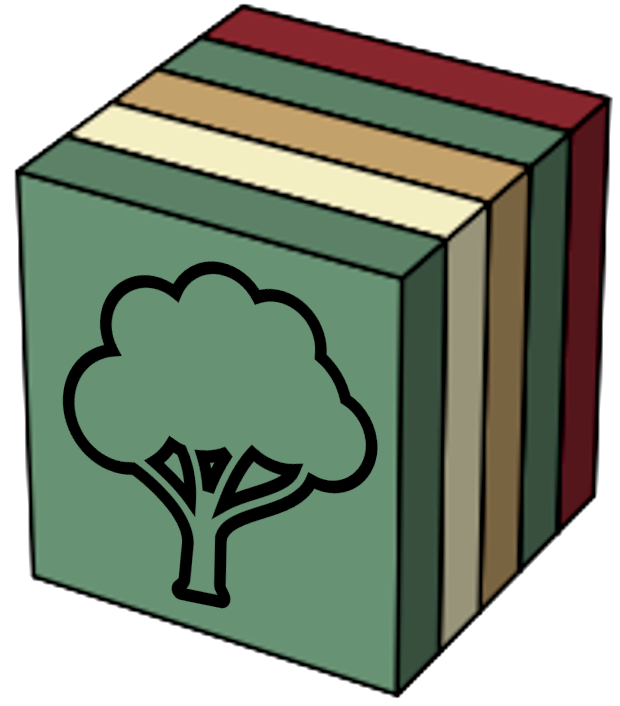}
        \subcaption{}
    \end{subfigure}
    \begin{subfigure}[b]{0.20\linewidth}
        \includegraphics[width=\linewidth]{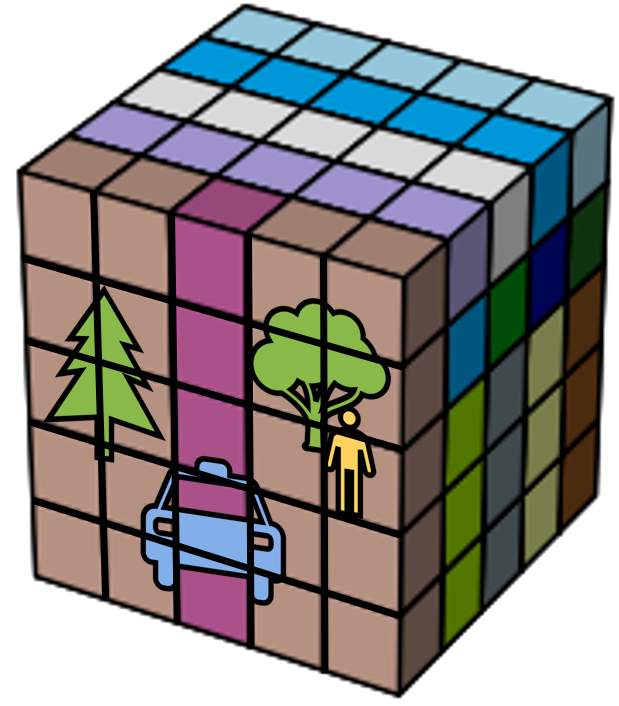}
        \subcaption{}
    \end{subfigure}
    \caption{
    Comparison of CTTA thresholding. Same colors indicate the same threshold. 
   (a) Fixed threshold in classification: a single value applied to the whole sample.
(b) Fixed threshold in segmentation: all pixels share the same threshold, causing pseudo-label errors.
(c) Batch-level adaptive threshold: class-wise but shared across instances.
(d) Ours: class thresholds are dynamically adjusted per instance for better segmentation adaptation.
    }
    \label{fig:threshold_}
\end{figure}

In this paper, we propose Continual Test-Time Instance and Class-wise Adaptation (CoTICA), a new framework that directly addresses these gaps for semantic segmentation. CoTICA introduces two complementary components. \textit{Instance-Class Adaptive Thresholding (ICAT)} moves beyond uniform or image-level thresholds by dynamically adjusting pseudo-label thresholds per class and per instance using pixel-wise class distributions. This enables fine-grained, temporally responsive pseudo labeling that better reflects evolving domain conditions. \textit{Instance-Class Weighted Loss (ICWL)} complements ICAT by prioritizing persistently challenging classes; it aggregates confidence across multiple augmentations and applies temporal smoothing to mitigate error accumulation in long-term adaptation (Figure~\ref{fig:threshold_}). Together, these modules provide a principled solution for pixel-level, class-aware, and temporally stable CTTA.

Our contributions are as follows:
\begin{itemize}
    \item We analyze the limitations of existing CTTA and dynamic thresholding approaches, showing their inability to adapt to pixel-level, class-dependent, and temporally evolving domain shifts in semantic segmentation.
    \item We propose CoTICA, which integrates Instance-Class Adaptive Thresholding and Instance-Class Weighted Loss to deliver fine-grained, class-aware, and temporally robust adaptation under continuously changing conditions.
    \item We establish new state-of-the-art performance across five CTTA and three TTA benchmarks, including challenging synthetic-to-real scenarios, demonstrating the effectiveness and novelty of our approach.
\end{itemize}

\section{Related Work}\label{related}

\subsection{Continual Test-Time Adaptation}
\label{subsec:tta}

Adapting models to changing data distributions without access to source data has become increasingly important. Early TTA methods initially focused on single-domain adaptation by minimizing entropy~\cite{wang2020tent, liang2020we, zhang2022memo} and employing self-supervision techniques~\cite{sun2020test, gandelsman2022test}, effectively tuning models for static target domains.
As real-world scenarios often involve continuously evolving domains, the field has shifted toward CTTA~\cite{wang2022continual, volpi2022road}. This new task aims to handle ongoing domain shifts, making it a more practical solution for dynamic environments.

Notably, CoTTA~\cite{wang2022continual} was among the first to introduce a teacher-student framework specifically designed for online continual adaptation. This method generates pseudo labels using the teacher model and updates it via a consistency loss, which has since influenced many subsequent studies. Building upon this, subsequent studies can be broadly categorized into efficiency-oriented and adaptation-oriented approaches. Efficiency-oriented methods mitigate catastrophic forgetting by freezing most model parameters—either updating only additional modules~\cite{song2023ecotta}, fine-tuning a small subset of weights~\cite{ni2024distribution}, or introducing learnable prompts at the input image level~\cite{gao2022visual, gan2023decorate, yang2024exploring}. These strategies effectively balance efficiency and robustness in non-stationary settings.
On the other hand, adaptation-oriented approaches explicitly address source–target discrepancies. For example, prototype-based methods~\cite{zhu2024reshaping, lee2025prototypical, dobler2023robust} optimize auxiliary objectives to maintain class prototypes across domains, while expert-based methods~\cite{liu2024vida, zhang2024decomposing} employ dedicated modules to handle domain-specific features and preserve domain-agnostic representations. 
Both directions demonstrate diverse yet effective pathways for achieving robust adaptation under continuous shifts.

Despite these advancements, previous studies have predominantly focused on classification tasks, with semantic segmentation often receiving only limited experimental attention. Although a few studies~\cite{ni2024distribution, yang2024exploring} have explored CTTA for segmentation, they often overlook the critical role of pseudo-labeling through augmentation, particularly in how thresholds are applied during the testing phase. Overall, existing methods~\cite{zhu2024reshaping, gan2023decorate, wang2022continual, ni2024distribution, yang2024exploring} use a fixed confidence threshold across all samples, which can lead to suboptimal performance under continuously shifting conditions. To tackle these challenges, we propose instance-class adaptive thresholding that dynamically reflects image and class-level variations.

\subsection{Adaptive Thresholding}
\label{subsec:at}
Pseudo labeling plays a crucial role in self-training for Continual Test-Time Adaptation (CTTA), particularly in scenarios where source data is unavailable. Since CTTA requires models to adapt to new domains solely based on target data, the quality of pseudo labels significantly impacts adaptation performance. However, in real-world settings, domain shifts and noisy predictions pose challenges for effective pseudo labeling, necessitating more adaptive strategies.  

Early approaches~\cite{wang2022continual, gao2022visual, song2023ecotta} have relied on fixed thresholding, where a prediction is assigned as a pseudo label if its confidence score surpasses a predefined threshold (\eg 0.9). While computationally efficient, these methods struggle to accommodate domain variations and uncertainty in predictions, often requiring extensive tuning for different environments. To enhance robustness, adaptive thresholding methods~\cite{guo2022class, zhang2021flexmatch} have been introduced. For instance, Li et al.~\cite{li2023robust} proposed evaluating multiple thresholds per sample to determine the optimal value, while Lee et al.~\cite{lee2023crowds} refined thresholds by comparing confidence scores before and after adaptation, filtering out unreliable pseudo labels. Furthermore, hybrid strategies that integrate batch-level global thresholds with class-specific local thresholds have been explored~\cite{yang2023exlore, wang2024continual, yang2024versatile}. However, these methods often rely on batch-level statistics, making them less effective for pixel-wise tasks such as semantic segmentation.  
For semantic segmentation, Mei et al.~\cite{mei2020instance} proposed a two-stage thresholding mechanism, which refines instance-specific thresholds using historical confidence trends via an Exponential Moving Average (EMA). While this method effectively balances historical and local information, it performs poorly under rapid domain shifts, where reliance on past trends can lead to outdated or misleading threshold adjustments.  

To address these limitations, we introduce ICAT (Instance and Class-wise Adaptive Thresholding), a novel method for semantic segmentation in CTTA. Unlike prior approaches that rely on historical trends, ICAT dynamically adjusts thresholds based on the confidence distribution within each instance while incorporating class-level adjustments. This allows pseudo-labeling to directly reflect domain shifts, improving segmentation accuracy in evolving environments. By leveraging both instance- and class-specific information, ICAT enhances pseudo-label precision and strengthens adaptation performance in CTTA.

\section{Method} 
\label{sec:method}

\subsection{Background}
\label{subsec:preliminary}

Continual Test-Time Adaptation (CTTA) aims to adapt a pre-trained model $f_{\theta_0}$ to a continuously evolving target domain $D = \{D_1, D_2, ..., D_k\}$ without requiring access to source data. This setting deals with \textit{continuous domain shifts}, where the target distribution gradually changes over time.  

A critical factor in CTTA is generating \textit{reliable pseudo labels}, as these labels directly influence the adaptation process. A widely adopted approach is the \textit{mean-teacher framework}~\cite{tarvainen2017mean}, which facilitates self-training through a teacher-student architecture. The student model $f_{\theta_t}$ produces predictions, while the teacher model $f_{\theta'_t}$, updated via an Exponential Moving Average (EMA), provides stable pseudo labels.  
To enhance robustness, prior CTTA methods~\cite{song2023ecotta, wang2024exploring, ni2024distribution, gan2023decorate, liu2024vida} often apply test-time augmentation~\cite{sun2020test, cohen2024simple} to the target sample, generating $N$ augmented views. The teacher aggregates these predictions to obtain an augmentation-averaged pseudo label:

\begin{equation}
    \label{eq:augmentation-averaged}
        \tilde{y}_{t}=\frac{1}{N}\sum_{i=1}^{N}f_{\theta'_t}(\text{aug}_i(x_t)).
\end{equation}

Pseudo labels are then selected based on model confidence, which serves as an indicator of domain shift severity~\cite{ni2024distribution, gan2023decorate, wang2024exploring}. A fixed threshold $\tau$ decides whether to use the original prediction $\hat{y}_t^{\prime}$ or the augmentation-averaged prediction $\tilde{y}_t$:

\begin{equation}
    \label{eq:fixed-threshold}
    y'_t = 
    \begin{cases}
        \hat{y}_t^{\prime}, & \text{if } \text{conf}(f_{\theta_t}(x_t)) \geq \tau, \\
        \tilde{y}_t, & \text{otherwise}.
    \end{cases}
\end{equation}

If the confidence is low, the model assumes a large domain shift and adopts $\tilde{y}_t$ to mitigate errors; otherwise, it retains $\hat{y}_t^{\prime}$. The student is updated with a consistency loss, while the teacher is updated via EMA.  

\begin{figure}[!t]
    \centering
    \includegraphics[width=0.86\columnwidth,keepaspectratio]{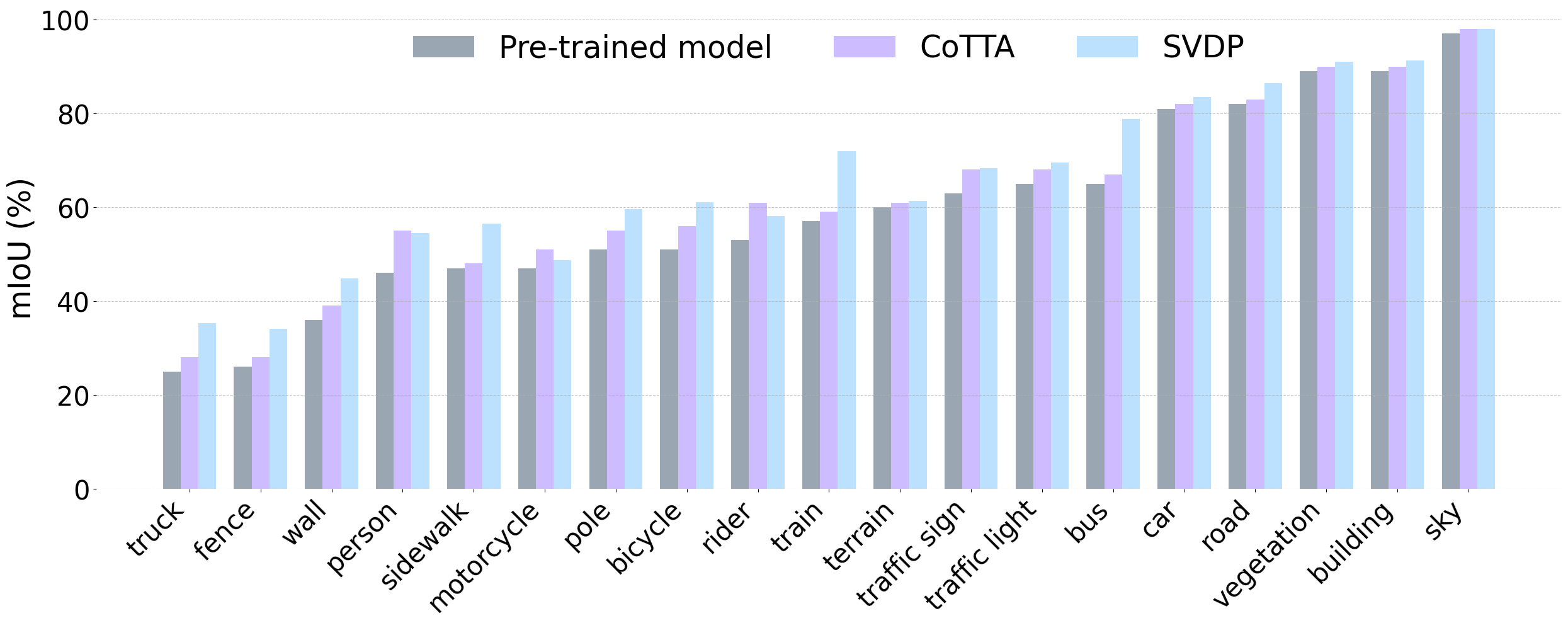}
        \caption{Class-wise mIoU (\%) comparison across different models under environmental changes on the ACDC dataset. The results highlight the need for adaptive strategies tailored to each class to improve segmentation performance.}
    \label{fig:model_test}
\end{figure}

However, this widely used strategy suffers from two limitations. First, using a fixed $\tau$ ignores varying domain conditions, often mislabeling classes that undergo significant shifts. Second, models treat all classes equally, even though some classes are consistently more affected by domain changes, leading to error accumulation over time. Figure~\ref{fig:model_test} highlights these issues by showing the performance inconsistency of existing approaches across different classes. These limitations motivate the need for a thresholding mechanism that adapts at the class and instance levels and a training objective that prioritizes more challenging classes.

\subsection{Instance and Class-wise Adaptation}
\label{subsec:method}

\begin{figure*}[t]
\centering
\includegraphics[width=1.0\textwidth,keepaspectratio]
{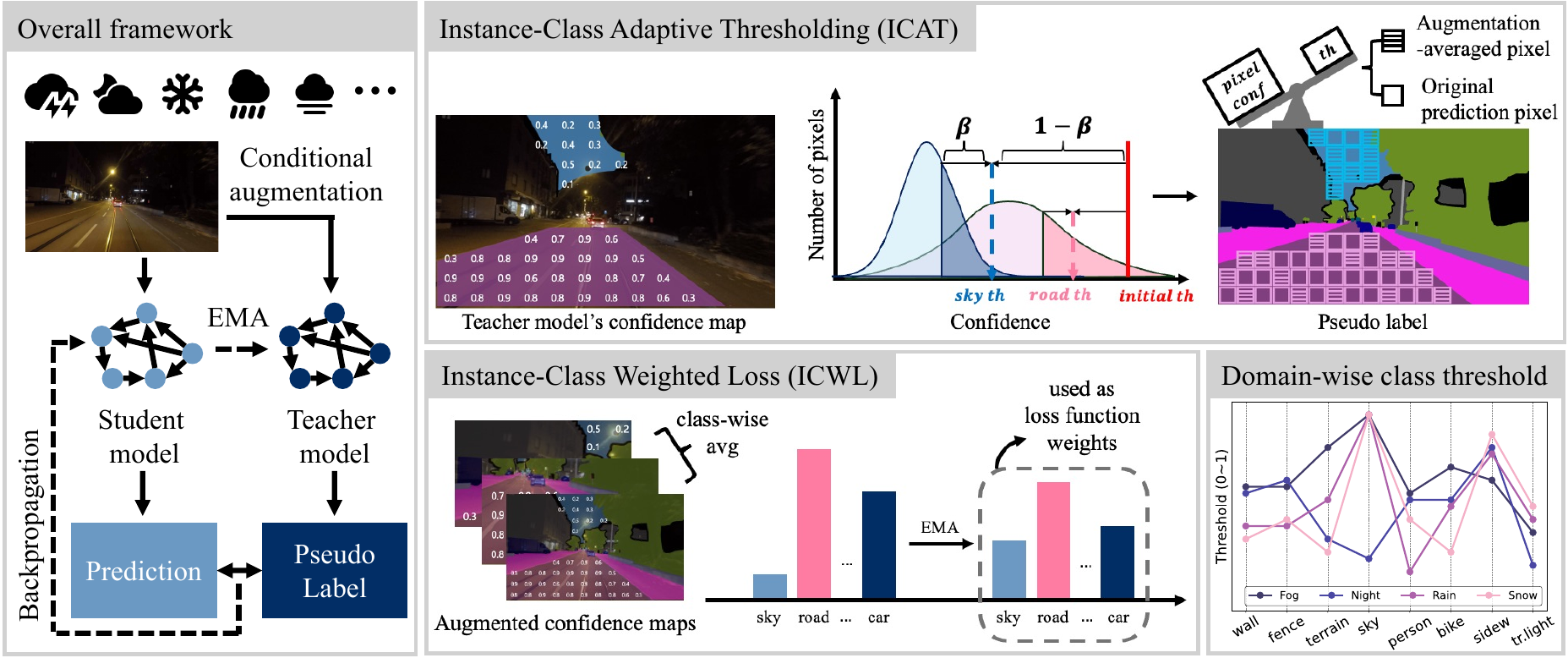}
\caption{
Overview of the proposed method. 1) A student-teacher model adapts to evolving domains using EMA updates. Conditional augmentation enhances pseudo-label robustness.
2) ICAT adjusts pseudo-label thresholds per instance and class using the teacher model’s confidence distribution. The process refines thresholds dynamically to improve adaptation.
3) ICWL computes class-wise importance weights from multiple augmented samples. These weights stabilize adaptation through moving-average updates.
4) Domain-wise Class Thresholds across different domains.
}
\label{figure_outline}
\end{figure*}

To overcome these challenges, we propose an {Instance- and Class-wise Adaptation} framework consisting of two complementary components: \textit{Instance-Class Adaptive Thresholding (ICAT)} and \textit{Instance-Class Weighted Loss (ICWL)}. Figure~\ref{figure_outline} illustrates the overall pipeline. ICAT dynamically adjusts pseudo-label thresholds for each class within an instance, ensuring reliable pseudo labeling under evolving shifts. ICWL builds on this by emphasizing adaptation for persistently hard classes through temporally smoothed difficulty estimation. By jointly optimizing ICAT and ICWL, our framework addresses both the unreliability of fixed thresholds and the lack of class-specific prioritization in existing CTTA methods.

\subsubsection{Instance-Class Adaptive Thresholding (ICAT)}
Confidence-based thresholding remains a natural choice for assessing domain gaps, as confidence scores reflect the model’s uncertainty. In semantic segmentation, each input $x_t$ yields pixel-wise confidence scores of size $H \times W$, where $H$ and $W$ are the image height and width. We group these scores by their predicted class $c$ to form a \textit{class-wise confidence distribution}:

\begin{equation}
    \label{eq:confidence-dist}
    P_{x_t}^c = \{ p_{h,w} \mid \arg\max f_{\theta_t}(x_t)_{h,w} = c \}, \quad p_{h,w} \in [0,1].
\end{equation}

Here, $P_{x_t}^c$ contains the confidence values of all pixels predicted as class $c$ in $x_t$. Using $P_{x_t}^c$, ICAT adaptively selects a threshold based on the $\alpha$-percentile of this distribution:

\begin{equation}
    \label{eq:icat-threshold}
    \phi(x_t^c) = P_{x_t}^c\big[\lfloor\alpha \cdot |P_{x_t}^c|\rfloor\big],
\end{equation}

where $|P_{x_t}^c|$ is the number of pixels predicted as class $c$, and $P_{x_t}^c[k]$ denotes the $k$-th value in the sorted confidence distribution. This percentile-based thresholding allows ICAT to adapt to each class’s difficulty: low-confidence classes receive lower thresholds, enabling aggressive adaptation, while high-confidence classes maintain stricter thresholds to preserve stability.  

To avoid instability from abrupt changes, ICAT blends this adaptive threshold $\phi(x_t^c)$ with the initial global threshold $\tau_0$:

\begin{equation}
    \label{eq:icat-final-threshold}
    \tau_t^c = \beta\tau_0 + (1-\beta)\phi(x_t^c),
\end{equation}

where $\beta\in[0,1]$ controls the balance between global and instance-class adaptivity. Final pseudo labels are then generated as:

\begin{equation}
    \label{eq:icat-pseudo}
    y'_t(h,w) = 
    \begin{cases}
        \hat{y}_t^{\prime}(h,w), & \text{if } \text{conf}(f_{\theta_t}(x_t)_{h,w}) \geq \tau_t^c, \\
        \tilde{y}_t(h,w), & \text{otherwise}.
    \end{cases}
\end{equation}

This adaptive mechanism ensures that pseudo labels reflect the current difficulty of each class within each instance, reducing the risk of erroneous updates and improving adaptation stability. Alg.~\ref{alg:cat_algorithm} provides a step-by-step summary of the ICAT procedure.

\begin{algorithm}[t]
\caption{Instance-Class Adaptive Thresholding}
\label{alg:cat_algorithm}
\small
\textbf{Input}: pre-trained model $\mathbf{f_{\theta_0}}$, teacher model $\mathbf{f_{\theta'_t}}$, target data $\mathbf{x}_{t}$  \\
\textbf{Parameter}: proportion $\alpha$, momentum $\beta$    \\
\textbf{Output}: target pseudo label $\mathbf{y'_t}$
\begin{algorithmic}[1]
\State \textbf{Initialize:} $\boldsymbol{\tau}_{0}=0.99$
\State $\mathbf{p}_{index}, \mathbf{p'}_{index} = \arg\max(\mathbf{f_{\theta_0}}(\mathbf{x}_{t})), \arg\max(\mathbf{f_{\theta'_t}}(\mathbf{x}_{t}))$
\State $\mathbf{p}_{value}, \mathbf{p'}_{value} = \max(\mathbf{f_{\theta_0}}(\mathbf{x}_{t})), \max(\mathbf{f_{\theta'_t}}(\mathbf{x}_{t}))$
\State $\mathbf{\tilde{y}_{t}} = \frac{1}{N}\sum_{i=1}^{N}\mathbf{f_{\theta'_t}}(\text{aug}_i(\mathbf{x}_{t}))$ \quad
\For{$c=1$ \textbf{to} $C$}
    \State $\mathbf{P}_{ \mathbf{x}_t}^c = \text{sort}(\mathbf{p'}_{value}[\mathbf{p'}_{index}=c], \text{descending})$
    \State $\boldsymbol{\phi}(\mathbf{x}_{t}^{c}) = \mathbf{P}_{ \mathbf{x}_t}^c[\alpha{\tau_0}\lvert{\mathbf{P}_{ \mathbf{x}_t}^c}\rvert]$ \quad 
\EndFor
\State $\boldsymbol{\tau}_{t}^{c} = \beta \boldsymbol{\tau} _{0}+(1-\beta)\boldsymbol{\phi}(\mathbf{x}_{t}^{c})$ \quad 
\State $\mathbf{M_t} = \mathbf{1}\{\mathbf{p}_{value}\geq\boldsymbol{\tau}_{t}^{c}\}$ 
\State $\mathbf{y'_t} = \mathbf{M_t} \cdot \mathbf{f_{\theta'_t}}(\mathbf{x}_{t}) + (1-\mathbf{M_t})\mathbf{\tilde{y}_{t}}$
\State \textbf{return} $\mathbf{y'_t}$
\end{algorithmic}
\end{algorithm}

\subsubsection{Instance-Class Weighted Loss (ICWL)}
While ICAT improves pseudo-label quality, some classes remain consistently harder to adapt to due to persistent domain shifts. To focus learning on these classes, we introduce \textit{Instance-Class Weighted Loss (ICWL)}.  

First, we estimate class-wise difficulty for the current sample by aggregating confidence values from $N$ augmented views:

\begin{equation}
    \label{eq:class-difficulty}
    \bar{y}_t^{\prime c} = \frac{1}{HW} \frac{1}{N} \sum_{i=1}^{N} \text{conf}(f_{\theta'_t}(\text{aug}_i(x_t)))^c.
\end{equation}

This reflects the model’s average confidence for class $c$ in $x_t$. To handle evolving domains, we stabilize these estimates over time using an EMA:

\begin{equation}
    \label{eq:ema-difficulty}
    \delta_t^c = \sigma\delta_{t-1}^c + (1-\sigma)\bar{y}_t^{\prime c}, \quad \delta_0^c = 1.
\end{equation}

Here, $\delta_{t-1}^c$ is the previous smoothed confidence for class $c$, $\sigma\in[0,1]$ controls the decay, and $\delta_0^c$ is initialized to 1, assuming full confidence at the start of adaptation. Lower $\delta_t^c$ indicates that class $c$ has become harder under the current domain shift.

Finally, we incorporate these smoothed class difficulties into the loss:

\begin{equation}
    \label{eq:icwl-loss}
    L_{\theta_t}(x_t) = -\sum_c (1-\delta_t^c)^\lambda y_t^{\prime c}\log(\hat{y}_t^c),
\end{equation}

where $y_t^{\prime c}$ is the refined pseudo label for class $c$, $\hat{y}_t^c$ is the student prediction, and $\lambda$ controls how strongly the model prioritizes low-confidence classes. By amplifying the contribution of hard classes, ICWL ensures that the model allocates more learning capacity where it is most needed.

Together, ICAT and ICWL form a unified framework that adaptively refines pseudo labels and directs learning toward persistently challenging classes, leading to robust and stable CTTA in semantic segmentation.

\section{Experiments}\label{experiments}

\subsection{Experimental Setup}
\label{subsec: experimental setup}

\begin{table*}[b]
    \centering
    \caption{Comparison of dataset splits, slass counts, and domain information}
    \resizebox{\textwidth}{!}{
    \begin{tabular}{lccccl}
        \toprule
        \textbf{Dataset} & \textbf{Train} & \textbf{Val} & \textbf{Test} & \textbf{Classes} & \textbf{Domain Information} \\
        \midrule
        ACDC         & --    & --    & 4,006  & 19 & fog, night, rain, snow \\
        Cityscapes   & 2,975             & 500                & 1,525  & 19 & daytime \\
        Cityscapes-C & 11,900             & 2,000                & 6,100  & 19 & bright, fog, frost, snow \\
        SHIFT        & --                & --               &   2.5M            & 19 & daytime$\rightarrow$night, clear$\rightarrow$foggy, clear$\rightarrow$rainy \\
        Dark Zurich  & 2,416 & --       & 151  & 19 & nighttime \\
        BDD100K      & 7,000            & 1,000             & 2,000 & 19 & daytime, nighttime, rainy, cloudy, etc. \\
        \bottomrule
    \end{tabular}
    }
    \label{tab:datasets_spec}
\end{table*}

\begin{table}[htbp]
    \centering
    \footnotesize
    \caption{Summary of the various domain shift scenarios.}
        \begin{tabular}{ll}
            \toprule
            
            \textbf{Source $\rightarrow$ Target} & \textbf{Detail} \\ 
            
            \midrule
            
             Cityscapes $\rightarrow$ ACDC & fog, night, rain, snow (CTTA, repeat 3 times)  \\
             Cityscapes $\rightarrow$ Cityscapes-C & brightness, fog, frost, snow (CTTA, repeat 10 times) \\
             SHIFT & discontinuous $\rightarrow$ continuous sequences (CTTA)  \\
             \midrule
             GTA5 $\rightarrow$ ACDC & \multirow{2}{*}{Synthetic-to-real (CTTA, repeat 3 times)}  \\
             GTA5 $\rightarrow$ Cityscapes-C \\
             \midrule
             GTA5 $\rightarrow$ DarkZurich & \multirow{3}{*}{Synthetic-to-real (TTA)} \\
             GTA5 $\rightarrow$ BDD100k  &  \\
             GTA5 $\rightarrow$ Cityscapes &  \\
            
             \bottomrule
        \end{tabular}
    \label{tab:settings}
\end{table}

\subsubsection{Datasets.} We used various semantic segmentation datasets to examine domain shifts. ACDC~\cite{sakaridis2021acdc} includes images under different weather conditions like fog and rain. Cityscapes-C~\cite{kamann2020benchmarking} was generated by applying algorithmic corruptions, such as noise, blur, weather effects, and digital distortions, to Cityscapes images~\cite{kamann2020benchmarking}. The SHIFT dataset~\cite{sun2022shift} simulates domain shifts with gradual transitions in environments, such as day to night. DarkZurich~\cite{sakaridis2019dark} features urban images taken during the day, twilight, and night, while BDD100k~\cite{yu2020bdd} offers 100,000 driving videos and images under diverse conditions. Table~\ref{tab:datasets_spec} summarizes the details of the datasets in our experiments.

\subsubsection{Task Settings.} We follow established protocols~\cite{yang2024exploring,song2023ecotta,sun2022shift} to ensure a fair comparison of our method across diverse task settings. First, we evaluate the source model, which was pre-trained on Cityscapes focusing on clear daytime scenes, on the ACDC benchmark under four different weather conditions: fog, rain, night, and snow. In this experiment, each evaluation is repeated three times to simulate real-world driving scenarios. The Cityscapes $\rightarrow$ Cityscapes-C benchmark tests robustness against various corruptions, with repeated 10 times. For SHIFT, focusing on natural domain shifts, the model is pre-trained on discontinuous synthetic sequences and tested on continuous synthetic sequences. We also introduce TTA and CTTA benchmarks for synthetic-to-real scenarios, with models pre-trained on GTA5. The target domains for TTA included DarkZurich, BDD100k, and Cityscapes, while for CTTA, they are ACDC and Cityscapes-C. To evaluate domain shifts effectively, the number of classes is standardized to match the pre-trained datasets. ACDC, Cityscapes, Cityscapes-C, DarkZurich, and BDD100k are evaluated using 19 classes, while the SHIFT dataset is evaluated using 14 classes. Table~\ref{tab:settings} summarizes the details of these task settings.

\subsubsection{Implementation Details.} For the SHIFT experiment, we used DeepLabV3~\cite{chen2017rethinking} with ResNet50~\cite{he2016deep}, setting the learning rate to 7.5e-6. For other experiments, we used Segformer-B5~\cite{xie2021segformer} with a learning rate of 6e-5. In all experiments, we employed the Adam optimizer~\cite{kingma2014adam} with ($\beta_1$, $\beta_2$) = (0.9, 0.999) and a batch size of 1. Additionally, we applied two augmentations to enhance the teacher model's robustness: \texttt{HorizontalFlip}, which flips the image horizontally with a 0.5 probability, and \texttt{RandomScale}, which resizes the image by a random factor from [0.5, 0.75, 1.0, 1.25, 1.5, 1.75, 2.0]. These augmentations generated 14 augmented samples as inputs to the teacher model. 

\begin{table*}[htbp]
    \setlength\tabcolsep{2.5pt}
    \caption{{Comparative analysis on Cityscapes to ACDC.} This table evaluates three iterations under four test conditions. `Mean' denotes the average mIoU(\%), and `Gain' represents the improvement in mIoU compared to the pre-trained model (`Source').}
    \centering
    \resizebox{\textwidth}{!}{
    \begin{tabular}{l|cc|cccc|cccc|cccc|c|c}
        \toprule
        \multicolumn{3}{l|}{Time}   & \multicolumn{12}{c|}{$t$ \makebox[10cm]{\rightarrowfill}} \\ \hline
        
        \multicolumn{3}{l|}{Round}  & \multicolumn{4}{c|}{1}    & \multicolumn{4}{c|}{2}     & \multicolumn{4}{c|}{3}   & \multirow{2}{*}{Mean}   & \multirow{2}{*}{Gain}  \\
        
        \multicolumn{1}{l}{Method} & ICWL & ICAT & Fog & Night & Rain & Snow & Fog & Night & Rain & Snow & Fog & Night & Rain & Snow & & \\ \hline
        
        \multicolumn{1}{l}{Source} & & &69.1&40.3&59.7&57.8 &69.1&40.3&59.7&57.8 &69.1&40.3&59.7&57.8 &56.7& 0.0\\ 
        
        \multicolumn{1}{l}{TENT~\cite{wang2020tent}} & & &69.0&40.2&60.1&57.3 &68.3&39.0&60.1&56.3 &67.5&37.8&59.6&55.0 &55.7&-1.0\\ 
        \multicolumn{1}{l}{CoTTA~\cite{wang2022continual}} & & &70.9&41.2&62.4&59.7 &70.9&41.1&62.6&59.7 &70.9&41.0&62.7&59.7 &58.6&+1.9\\ 
        \multicolumn{1}{l}{DePT~\cite{gao2022visual}} & & &71.0&40.8&58.2&56.8 &68.2&40.0&55.4&53.7 &66.4&38.0&47.3&47.2 &53.4&-3.3\\
        \multicolumn{1}{l}{VDP~\cite{gan2023decorate}} & & &70.5&41.1&62.1&59.5 &70.4&41.1&62.2&59.4 &70.4&41.0&62.2&59.4 &58.2&+1.5\\
        \multicolumn{1}{l}{DAT~\cite{ni2024distribution}} & & &71.7&44.4&65.4&62.9 &71.6&\textbf{45.2}&63.7&63.3 &70.6&44.2&63.0&62.8 &60.8&+4.1\\
        \multicolumn{1}{l}{OBAO~\cite{zhu2024reshaping}} & & &71.2 &42.3 &64.9 &62.0 &72.6 &43.2 &66.3 &63.2 &72.8 &43.8 &66.5 &63.2 &61.0 &+4.3\\
        \multicolumn{1}{l}{SVDP~\cite{yang2024exploring}} & & &72.1&44.0&65.2&63.0 &72.2&44.5&65.9&63.5 &72.1&44.2&65.6&63.6 &61.3&+4.6\\
        \multicolumn{1}{l}{ADMA~\cite{liu2024continual}} & & &71.9&\textbf{44.6}&\textbf{67.4}&63.2 &71.7&44.9&66.5&63.1 &72.3&\textbf{45.4}&67.1&63.1 &61.8&+5.1\\
        \multicolumn{1}{l}{ViDA~\cite{liu2024vida}} & & &71.6&43.2&66.0&63.4 &73.2&44.5&67.0&63.9 &73.2&44.6&67.2&64.2 &61.9&+5.2\\
        \multicolumn{1}{l}{MoASE~\cite{zhang2024decomposing}} & & &72.4 &44.5 &66.4 &63.8 &73.0 &45.1 &67.5 &63.5 &73.5 &44.5 &67.4 &63.5 &62.2 &+5.5\\
        \rowcolor{blue!10} 
        \multicolumn{1}{l}{\textbf{Ours}} & \checkmark &  &72.8&43.6&63.9&63.0 &73.9&44.3&64.4&63.4 &74.0&44.9&64.7&63.3 &61.3&+4.6\\
        \rowcolor{blue!10} 
        \multicolumn{1}{l}{\textbf{Ours}} &  & \checkmark &73.9&43.8&66.3&64.5 &74.9&43.8&67.0&64.9 &75.1&43.2&67.1&64.8 &62.4&+5.7\\
        \rowcolor{blue!10} 
        \multicolumn{1}{l}{\textbf{Ours}} & \checkmark & \checkmark &\textbf{74.4}&43.5&66.5&\textbf{65.2} &\textbf{75.3}&44.2&\textbf{67.4}&\textbf{65.5} &\textbf{75.4}&44.4&\textbf{67.3}&\textbf{65.3} &\textbf{62.9}&\textbf{+6.2}\\
        \bottomrule
    \end{tabular}
    }
    \label{tab:CTTA}
\end{table*}

The hyperparameters used in the proposed method were set as follows: For ICAT, the initial threshold $\tau_0$ was set to 0.99 for the Segformer-B5 backbone and 0.8 for the DeepLabV3 backbone, with $\alpha = 0.2$ and $\beta = 0.9$ applied uniformly across all backbones. For ICWL, the scaling factor $\lambda$ and decay factor $\sigma$ were set to 3 and 0.999, respectively. All experiments were conducted using an NVIDIA RTX A6000 GPU.

\subsection{Comparison Results}

\subsubsection{Cityscapes to ACDC.}
Table~\ref{tab:CTTA} presents the quantitative results on the ACDC dataset. The \textit{Source} approach, which performs inference without adaptation, achieves 56.7\% mIoU, emphasizing the need for domain adaptation in diverse environments.  
TTA-based methods such as TENT and DePT perform well initially but degrade over time in CTTA due to accumulated errors. CTTA methods like CoTTA and VDP alleviate this issue but remain limited, as they primarily target classification rather than segmentation.  
Recent segmentation-focused CTTA approaches~\cite{ni2024distribution, yang2024exploring, liu2024continual, liu2024vida} show notable improvements. Our method achieves 62.9\% mIoU, outperforming \textit{Source} by 6.2\%. This demonstrates the effectiveness of incorporating instance- and class-wise thresholding and weighting for robust CTTA in semantic segmentation.

\subsubsection{Cityscapes to Cityscapes-C.}
\begin{table*}[]
    \setlength\tabcolsep{2.5pt}

    \caption{{Comparative analysis on Cityscapes to Cityscapes-C.} We evaluate long-term adaptation by performing 10 iterations for each of the four conditions.}
    \centering
    \resizebox{\textwidth}{!}{
    \begin{tabular}{l|cccc|cccc|cccc|cccc|c}
        \toprule
        Time & \multicolumn{16}{c|}{$t\xrightarrow{\hspace*{14cm}}$}  \\   
        \midrule
        Round &  \multicolumn{4}{c|}{1} & \multicolumn{4}{c|}{4} & \multicolumn{4}{c|}{7} & \multicolumn{4}{c|}{10} & \multirow{2}{*}{Mean}  \\ 
        Condition & Brig. & Fog &Fros. & Snow & Brig. & Fog &Fros. & Snow & Brig. & Fog &Fros. & Snow & Brig. & Fog &Fros. & Snow & \\ 
        \midrule

        Backbone & \multicolumn{17}{c}{DeepLabV3+} \\

        \midrule
        
        Source &  60.4 & 54.3 & 30.0 & 4.1  & 60.4 & 54.3 & 30.0 & 4.1 & 60.4 & 54.3 & 30.0 & 4.1 & 60.4 & 54.3 & 30.0 & 4.1 & 37.2 \\
        BN Adapt &   69.1 & 61.0 & 44.8 & 39.1 & 69.1 & 61.0 & 44.8 & 39.1 & 69.1 & 61.0 & 44.8 & 39.1 & 69.1 & 61.0 & 44.8 & 39.1 & 53.6  \\
        TENT~\cite{wang2020tent} &  70.1 & 62.1 & 46.1 & 40.2    & 62.2 & 53.7 & 44.4 & 37.9     & 50.0 & 41.5 & 31.6 & 26.6       & 39.2 & 32.6 & 25.3 & 22.4 & 42.9 \\ 
        EcoTTA~\cite{song2023ecotta} &  70.2 & 62.4 & \textbf{46.3} & 41.9     & 70.0 & 62.8 & 46.5 & 42.2    & 70.0 & 62.8 & 46.5 & 42.1     & 70.1 & 62.8 & \textbf{46.6} & 42.2 & 55.3   \\ 
        \midrule

        Backbone & \multicolumn{17}{c}{SegFormer-B5} \\

        \midrule
        
        Source 
        &75.7&67.3&33.0&38.6  &75.7&67.3&33.0&38.6 &75.7&67.3&33.0&38.6  &75.7&67.3&33.0&38.6   &53.6 \\ 
        TENT~\cite{wang2020tent} & 75.6&67.1&32.9&37.4   &72.6&62.7&30.8&33.7     &69.3&59.1&28.7&31.5      &66.5&55.9&27.3&30.2   & 48.8 \\ 
        CoTTA~\cite{wang2022continual} & 78.1&70.4&34.6&39.1   &78.1&70.4&34.6&39.0  &\textbf{78.1}&70.3&34.5&39.0	&\textbf{78.0}&70.4&34.4&38.9     &55.5 \\ 
        DAT~\cite{ni2024distribution}  
        &78.5&73.5&44.8&49.0   &74.6&68.9&\textbf{47.4}&50.5   &73.2&67.5&45.4&48.9	&72.0&66.4&43.5&47.5    &59.6 \\ 
        SVDP~\cite{yang2024exploring} 
        &78.3&72.9&43.7&\textbf{52.1}   &73.7&68.5&44.7&\textbf{51.8}   &72.8&68.0&\textbf{46.8}&\textbf{52.7}	&72.9&66.9&46.2&\textbf{51.8}     &59.9 \\
        \rowcolor{blue!10} %
        \textbf{Ours} &  \textbf{79.2} & \textbf{73.9} & 39.8 & 47.5   & \textbf{78.7} & \textbf{73.3} & 42.1 & 50.6   & 77.7 & \textbf{72.5} & 40.6 & 48.6     & 77.6 & \textbf{72.2} & 40.0 & 48.4 & \textbf{60.3}   \\ 
        \bottomrule
    \end{tabular}
    }
    \label{tab:cityscapes}
\end{table*}

Table~\ref{tab:cityscapes} summarizes results on Cityscapes-C, evaluating long-term adaptation under corruption-induced shifts. Existing methods often suffer from severe error accumulation, resulting in significant performance drops over time. For example, SVDP~\cite{yang2024exploring} loses 5\% mIoU in the fog domain by the final iteration.  
EcoTTA~\cite{song2023ecotta} and CoTTA~\cite{wang2022continual} maintain relatively stable results but still face limitations in pseudo-label quality. Our method sustains high performance across all corruption types by dynamically refining pseudo labels per instance and class, effectively preventing error propagation during continual adaptation.

\subsubsection{SHIFT.}
\begin{table*}[]
    \caption{Comparative analysis on SHIFT's gradual environmental transitions.} %
    \centering
    \resizebox{0.8\textwidth}{!}{
        \begin{tabular}{l|rr|rr|rr|rr}
            \toprule
            Scenario  & \multicolumn{2}{c|}{Daytime $\rightarrow$ Night} & \multicolumn{2}{c|}{Clear $\rightarrow$ Foggy} & \multicolumn{2}{c|}{Clear $\rightarrow$ Rainy} & \multicolumn{2}{c}{Mean} \\        
            \midrule
            Method & mIoU & Acc & mIoU & Acc & mIoU & Acc & mIoU & Acc \\
            \midrule
            Source  & 55.5 & 65.5 & 44.2 & 56.3 & 46.7 & 60.1 & 49.4 & 61.1\\ 
            TENT~\cite{wang2020tent}  & 55.3 & 62.6 & 43.1 & 52.2 & \textbf{49.7} & 58.6 & 50.0 & 58.3\\
            CoTTA~\cite{wang2022continual} & 57.4 & 66.8 & 44.4 & 55.4 & 48.6 & 61.5 & 50.8 & 61.8  \\
            \rowcolor{blue!10} %
            \textbf{Ours} & \textbf{57.6} & \textbf{66.9} & \textbf{45.2} & \textbf{56.4}& 48.9 & \textbf{61.7} & \textbf{51.3} & \textbf{62.2}\\
            \bottomrule
        \end{tabular}
        }
    
    \label{tab:shift}
\end{table*}

Table~\ref{tab:shift} shows results on the SHIFT dataset, which simulates gradual transitions (e.g., day-to-night, clear-to-fog/rainy). This incremental change poses greater challenges than single-step adaptation.  
TENT~\cite{wang2020tent} often underperforms the Source baseline due to unstable updates. CoTTA~\cite{wang2022continual} mitigates forgetting but still struggles with progressive shifts. In contrast, CoTICA consistently delivers state-of-the-art performance across nearly all scenarios, demonstrating strong adaptability in non-stationary environments.

\subsubsection{Synthetic to Real.}
\begin{table*}[]
    \setlength\tabcolsep{2.5pt}
    \caption{Comparative analysis of synthetic-to-real adaptation using GTA5 as the source dataset. Results for three TTA scenarios (DarkZurich (DZ), BDD100k (BDD), and Cityscapes (City)) and two CTTA scenarios (Cityscapes-C and ACDC).} 
    \centering
    \resizebox{0.8\textwidth}{!}{
        \begin{tabular}{l|ccc|cccccccc}
            \toprule
            Scenario & \multicolumn{3}{c|}{TTA} & \multicolumn{8}{c}{CTTA}\\ \hline
            \multirow{2}{*}{Method} & \multicolumn{1}{c|}{\multirow{2}{*}{DZ}} & \multicolumn{1}{c|}{\multirow{2}{*}{BDD}} & \multirow{2}{*}{City} & \multicolumn{4}{c|}{Cityscapes-C} & \multicolumn{4}{c}{ACDC}\\ 
            & \multicolumn{1}{c|}{} & \multicolumn{1}{c|}{} & & 1 & 2 & 3 & \multicolumn{1}{c|}{Mean} & 1 & 2 & 3 & Mean \\ \hline
            Source & \multicolumn{1}{c|}{18.5} & \multicolumn{1}{c|}{41.2} & 46.8 & 28.6 & 28.6 & 28.6 & \multicolumn{1}{c|}{28.6} & 37.7 & 37.7 & 37.7 & 37.7 \\
            TENT~\cite{wang2020tent} & \multicolumn{1}{c|}{18.5} & \multicolumn{1}{c|}{41.4} & 46.8 & 29.0 & 29.2 & 28.1 & \multicolumn{1}{c|}{28.8} & 37.6 & 37.6 & 36.1 & 37.1 \\
            CoTTA~\cite{wang2022continual} & \multicolumn{1}{c|}{17.8} & \multicolumn{1}{c|}{40.5} & 48.0 & 28.1 & 27.6 & 27.4 & \multicolumn{1}{c|}{27.7} & 36.6 & 35.7 & 35.1 & 35.8 \\
            DAT~\cite{ni2024distribution} & \multicolumn{1}{c|}{17.9} & \multicolumn{1}{c|}{41.8} & 48.7 & 31.7 & 26.1 & 25.0 & \multicolumn{1}{c|}{27.5} & 37.7 & 36.9 & 36.0 & 36.9 \\
            \rowcolor{blue!10} %
            \textbf{Ours} & \multicolumn{1}{c|}{\textbf{19.6}} & \multicolumn{1}{c|}{\textbf{44.5}} & \textbf{50.5} & \textbf{32.7} & \textbf{32.8} & \textbf{32.4} & \multicolumn{1}{c|}{\textbf{32.7}} & \textbf{39.9} & \textbf{39.6} & \textbf{39.4} & \textbf{39.6} \\
            \bottomrule
        \end{tabular}
    }
    \label{tab:sim2real}
\end{table*}

Table~\ref{tab:sim2real} compares methods in synthetic-to-real adaptation using GTA5 as the source. This setting involves large domain gaps compounded by variations in weather and illumination across datasets such as DarkZurich, BDD100k, Cityscapes-C, and ACDC.  
Pseudo label–based methods like CoTTA and DAT, while effective elsewhere, degrade here due to static thresholds. Non-pseudo-label methods (Source, TENT) remain more stable but lack adaptability. By dynamically adjusting thresholds per instance and class, our approach consistently achieves the highest accuracy, demonstrating its strength in bridging synthetic-to-real gaps.

\begin{figure*}[]
\centering
\includegraphics[width=1.0\textwidth,keepaspectratio]{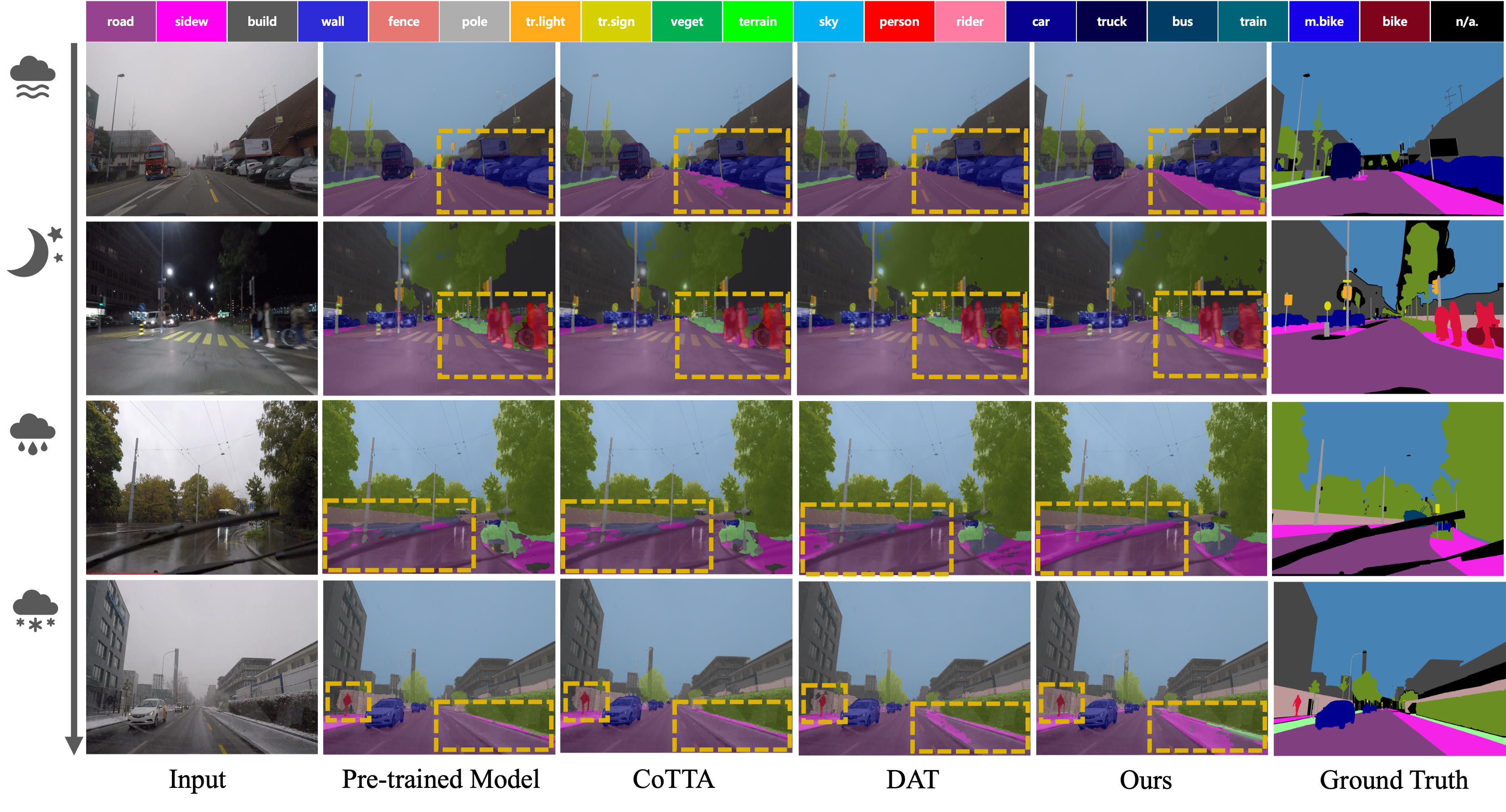}
\caption{
{Comparison with qualitative results.} The yellow bounding box highlights areas with qualitative performance improvements, pinpointing regions that were previously challenging for previous methods.
} 
\label{qualitative}
\end{figure*}

\subsubsection{Qualitative Results.}
Figure~\ref{qualitative} illustrates the qualitative improvements of our method on ACDC. It enhances segmentation for major objects like sidewalks while improving finer details, such as fences, which are often missed by previous methods.
Notably, the yellow bounding boxes highlight challenging regions where prior approaches struggle, particularly under adverse weather conditions such as nighttime, rain, and snow. Compared to previous methods, our method produces more stable and consistent predictions, reducing artifacts and misclassifications in low-visibility scenarios. These refinements lead to higher-quality pseudo labels, which in turn contribute to better long-term adaptation in dynamic environments.
By effectively capturing both large structural elements and fine-grained details, our method ensures more reliable segmentation across diverse domain-shift conditions, making it particularly valuable for real-world applications such as autonomous driving.

\subsection{Ablation Study}

\begin{figure}[]
    \centering
    \begin{subfigure}[b]{0.42\linewidth}
        \centering
        \includegraphics[width=\linewidth]{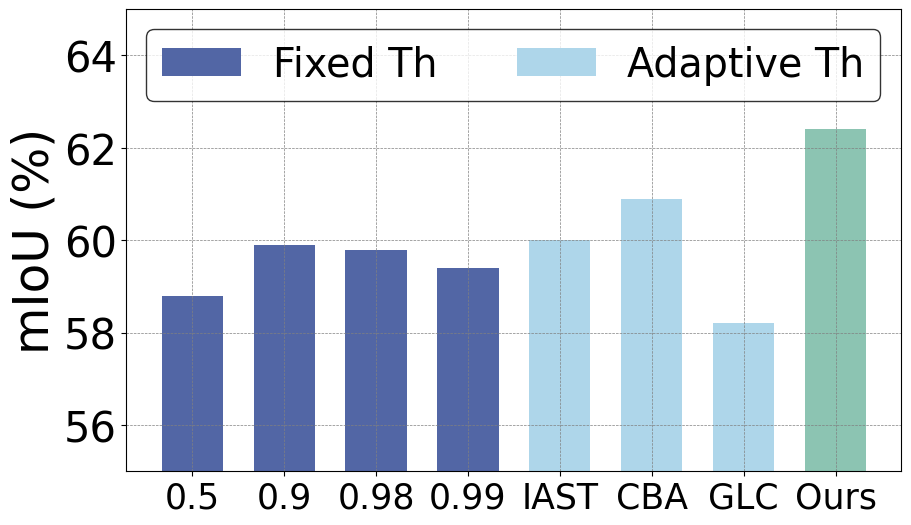}
        \caption{}
        \label{compare_cat}
    \end{subfigure}
    \begin{subfigure}[b]{0.34\linewidth}
        \centering
        \includegraphics[width=\linewidth]{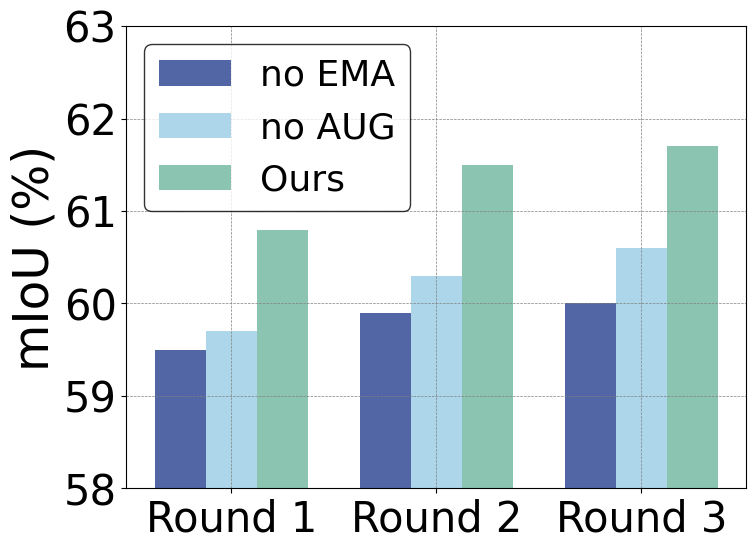}
        \caption{}
        \label{compare_icwl}
    \end{subfigure}
    \caption{Analysis of the proposed key modules (a) Performance comparison based on threshold strategies used for augmentation
(b) Ablation study of ICWL}
    \label{fig:combined}
\end{figure}

\subsubsection{Thresholding strategy comparison.}

Figure~\ref{compare_cat} illustrates the improvements achieved by ICAT compared to existing thresholding methods. We evaluated ICAT against various fixed and adaptive thresholding approaches. For fixed thresholds, we selected values around 0.98, the average per-class threshold obtained by ICAT. Results showed that fixed thresholds generally led to lower performance, with a significant drop at 0.5, the farthest from the average. This highlights the model’s sensitivity to threshold selection, underscoring the need for adaptive strategies.

Next, we compared ICAT with IAST\cite{mei2020instance}, CBA\cite{lee2023crowds}, and GLC~\cite{yang2023exlore, wang2024continual, yang2024versatile}. IAST stabilizes threshold adjustments using EMA but struggles with abrupt domain shifts, limiting its responsiveness in dynamic environments. CBA filters noisy predictions based on confidence differences between pre- and post-adaptation models but lacks the granularity needed for pixel-level segmentation. GLC refines pseudo labels using global and class-specific thresholds but remains batch-level, making it less effective for segmentation tasks requiring precise pixel-wise adaptation.
In contrast, ICAT eliminates reliance on historical data and dynamically adjusts thresholds at both instance and class levels. By adapting in real time based on the current confidence distribution, ICAT effectively handles sudden domain shifts and diverse data distributions.

\begin{figure}[]
    \centering
    \begin{subfigure}[b]{0.43\columnwidth}
        \includegraphics[width=\linewidth]{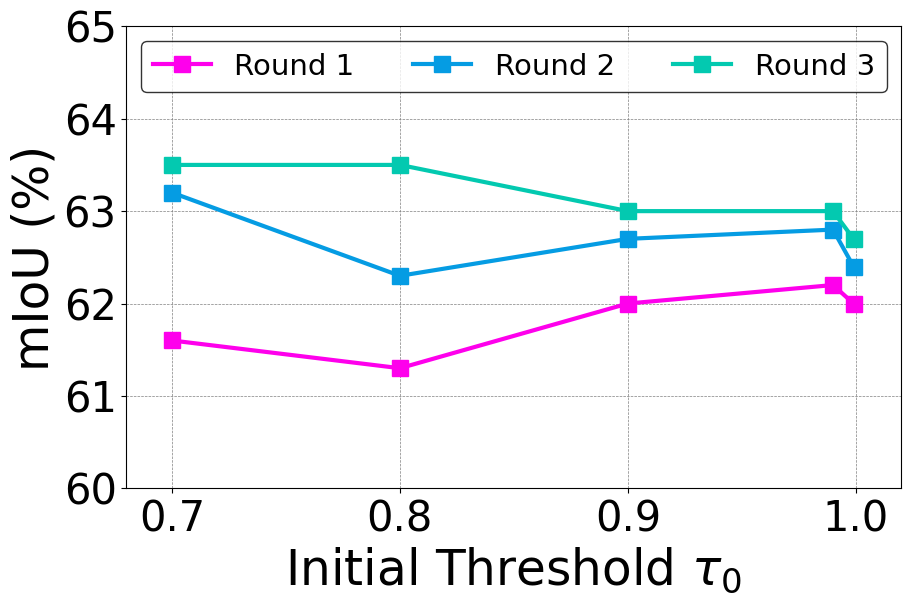}
        \caption{}
    \label{cat_sensitivity}
    \end{subfigure}
    \begin{subfigure}[b]{0.43\columnwidth}
        \includegraphics[width=\linewidth]{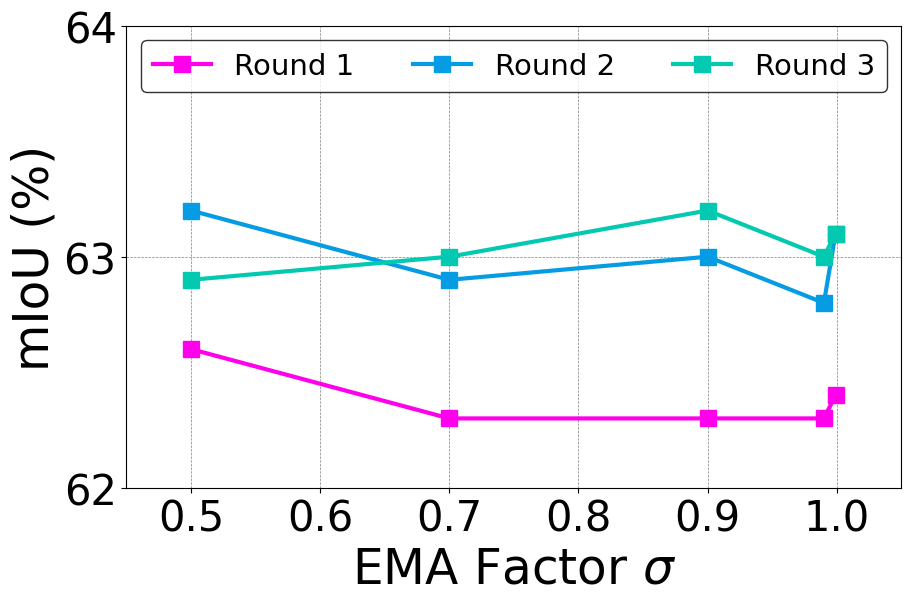}
        \caption{}
    \label{hwl_sensitivity}
    \end{subfigure}
    \caption{
    Sensitivity analysis of (a) ICAT and (b) ICWL.
        }
\end{figure}

\subsubsection{Ablation Study of ICWL.}

We conducted an ablation study on the ACDC dataset to analyze the impact of data augmentation (AUG) and the exponential moving average (EMA) in our ICWL. AUG enhances sample diversity to mitigate class imbalance, while EMA refines learning weights continuously, stabilizing adaptation over time.
As shown in Figure~\ref{compare_icwl}, the `no EMA', which relies only on current samples, shows minimal performance gains across rounds, indicating instability in weight estimation. The `no AUG' provides moderate improvements but lacks the sample diversity needed for accurate weighting, limiting its effectiveness. In contrast, our full method, which combines both components, achieves the highest and most stable performance, demonstrating the importance of continual weight refinement and diverse sampling for robust CTTA segmentation.

\subsection{In-depth Analysis}

\subsubsection{Sensitivity Analysis.}

We conducted a hyperparameter sensitivity analysis. The experiments were performed over three iterations for each environmental transition (fog $\rightarrow$ night $\rightarrow$ rain $\rightarrow$ snow) in the ACDC dataset, with performance evaluated using the mIoU metric.
For ICAT, we varied the initial threshold $\tau_0$ across values of 0.7, 0.8, 0.9, 0.99, and 0.999 while applying ICWL simultaneously. As shown in Figure~\ref{cat_sensitivity}, $\tau_0$ = 0.99 achieved the highest mIoU in the first round. However, lower thresholds of 0.7 and 0.8 yielded better results in subsequent rounds, with the best overall performance observed in the final round. This indicates that higher initial thresholds do not always lead to improved outcomes, suggesting a need for balance in setting $\tau_0$.
For ICWL, we examined the effect of the EMA factor $\sigma$, which influences the impact of previous class-wise learning weights $\delta_{t-1}^c$ on current performance. Figure~\ref{hwl_sensitivity} shows that $\sigma$ = 0.5 produced the highest mIoU in rounds 1 and 2 but declined sharply in round 3. In contrast, $\sigma$ values of 0.9 and 0.999 provided more consistent performance across all rounds. This suggests that more frequent reference to previous class-specific weights benefits continuous learning, leading to more stable improvements over time.

\begin{figure}[!tb]
    \centering
    \includegraphics[width=0.8\columnwidth,keepaspectratio]{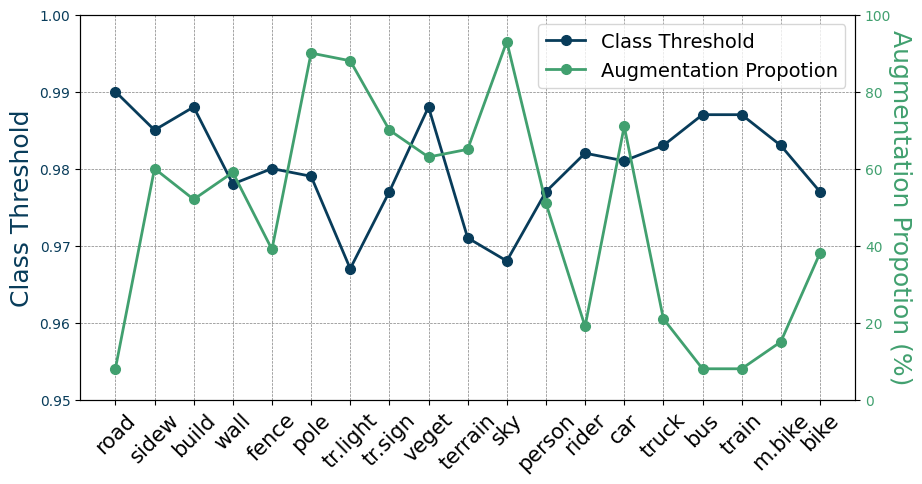}
        \caption{
            The correlation between the usage ratio of augmentation and ICAT values. This inverse relationship indicates that ICAT effectively handles difficult classes by leveraging augmented averages.
        }
    \label{aug_cat}
\end{figure}

\subsubsection{Relationship between Threshold and Augmentation}
Figure~\ref{aug_cat} illustrates the inverse relationship between class thresholds and augmentation usage rates, demonstrating how ICAT optimizes pseudo-labeling. ICAT dynamically adjusts thresholds based on each class’s confidence distribution, assigning lower thresholds to low-confidence classes while maintaining relatively high absolute thresholds across all classes.
As described in the main paper, ICAT determines whether augmentation is applied by comparing pixel confidence with the class threshold. Low-confidence classes typically struggle due to insufficient training data or significant distribution shifts, making predictions less reliable. To address this, ICAT increases augmentation for these classes, leveraging augmented averages to improve pseudo-label quality. Although these classes have relatively lower thresholds, their absolute values remain high, making threshold exceedance rare.
Consequently, ICAT introduces more augmented predictions for hard-to-predict classes, helping the model adapt to diverse distributions. In contrast, high-confidence classes with higher thresholds receive fewer augmentations, as they require less adaptation. This adaptive mechanism enables CoTICA to generate robust, class-aware pseudo labels, ensuring balanced adaptation across varying class difficulties.

\begin{figure*}[!tb]
    \centering
    \includegraphics[width=1.0\textwidth,keepaspectratio]{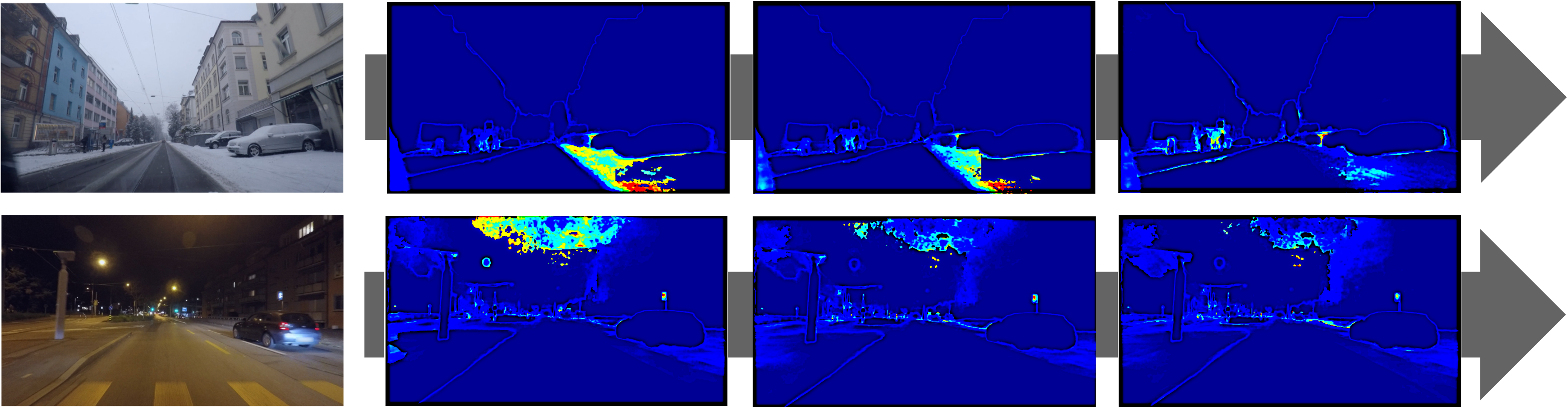}
        \caption{
            Visualization of the focus of our ICWL across different weather conditions. ICWL shows a dynamic focus on challenging regions, with a significant reduction in loss from early to late adaptation epochs.
            } 
    \label{heatmap}
\end{figure*}

\subsubsection{Heatmap Visualization}

To demonstrate the effectiveness of our proposed ICWL, we conducted heatmap visualizations.
{Figure~\ref{heatmap} illustrates how the model’s focus evolves during the adaptation process, from the early to middle and late stages.}
In the night domain, high-loss values are particularly observed in the sky regions during early epochs, while high-loss values appear on the sidewalks in the snow domain. These high-loss areas indicate that the model is focusing its learning on challenging classes. As the adaptation progresses to the middle and late epochs with ICWL applied, the initially high-loss values in the sky and sidewalk pixels decrease significantly.
{This reduction in loss demonstrates ICWL's ability to maintain focus on challenging classes while gradually improving its adaptation and overall learning performance.}

\subsection{Limitations and Future Directions.} 
Despite the strong performance of our method, efficiency remains a challenge. Like previous CTTA approaches~\cite{wang2022continual, dobler2023robust, brahma2023probabilistic}, which often struggle to balance performance and computational efficiency~\cite{ni2024distribution, song2023ecotta}, our method also incurs additional resource demands. The need to compute and maintain adaptive thresholds and loss weights per class increases both memory usage and computational cost.
To mitigate this, we explored parameter-efficient fine-tuning techniques like LoRA~\cite{hu2021lora}, but this led to performance degradation, making it unsuitable for handling continuous domain shifts effectively. As a result, we prioritized maintaining robust adaptation capabilities over reducing computational overhead.
Future work will focus on optimizing efficiency without sacrificing adaptation performance, making the approach more practical for real-time and resource-constrained applications across diverse environments.

\section{Conclusion}
In this paper, we introduced Continual Test-time Instance and Class-wise Adaptation (CoTICA), a robust framework designed to tackle the challenges of Continual Test-Time Adaptation (CTTA) in semantic segmentation. CoTICA effectively adapts to evolving domain shifts by leveraging Instance-Class Adaptive Thresholding (ICAT) and Instance-Class Weighted Loss (ICWL), enabling instance-specific and class-wise adjustments for more precise adaptation.
ICAT dynamically adjusts pseudo-label thresholds based on the class distribution within each image, generating accurate pixel-level pseudo labels tailored to dynamic environments. This ensures fine-grained adaptation in segmentation tasks where domain shifts vary across classes and instances. ICWL further enhances adaptation by dynamically updating class-specific learning weights, balancing learning between low- and high-confidence classes. This strategy prevents error accumulation and stabilizes adaptation across different conditions.
By integrating instance- and class-aware adaptation, CoTICA improves model robustness and ensures consistent performance across diverse and evolving domains. Experimental results demonstrate that CoTICA outperforms existing CTTA methods and alternative thresholding strategies across eight TTA and CTTA scenarios, setting a new benchmark for continual adaptation.


\newpage







\begin{thebibliography}{00}


\bibitem[Siqi(2023)]{fan2023conservative}
  Siqi Fan, Fenghua Zhu, Zunlei Feng, Yisheng Lv, Mingli Song, and Fei-Yue Wang,
  Conservative-progressive collaborative learning for semi-supervised semantic segmentation,
  In \textit{{IEEE Transactions on Image Processing}}, vol. 32, pages 6183--6194, 2023.

\bibitem[Lukas(2022)]{hoyer2022hrda}
  Lukas Hoyer, Dengxin Dai, and Luc Van Gool,
  HRDA: Context-aware high-resolution domain-adaptive semantic segmentation,
  In \textit{{Proceedings of the European Conference on Computer Vision (ECCV)}}, pages 372--391, 2022.



\bibitem[Haotian(2024)]{liu2024visual}
  Haotian Liu,Chunyuan Li, Qingyang Wu, and Yong Jae Lee,
  Visual instruction tuning,
  In \textit{{Proceedings of Neural Information Processing Systems (NeurIPS)}}, vol. 36, 2024.

\bibitem[Dian(2022)]{chen2022contrastive}
  Dian Chen,Dequan Wang, Trevor Darrell, and Sayna Ebrahimi,
  Contrastive test-time adaptation,
  In \textit{{Proceedings of the IEEE/CVF Conference on Computer Vision and Pattern Recognition (CVPR)}}, pages 295--305, 2022.

\bibitem[Shuaicheng(2022)]{niu2022efficient}
  Shuaicheng Niu,Jiaxiang Wu, Yifan Zhang, Yaofo Chen, Shijian Zheng, Peilin Zhao, and Mingkui Tan,
  Efficient test-time model adaptation without forgetting,
  In \textit{{Proceedings of International Conference on Machine Learning (ICML)}}, pages 16888--16905, 2022.

\bibitem[Malik(2022)]{boudiaf2022parameter}
  Malik Boudiaf,Romain Mueller, Ismail Ben Ayed, and Luca Bertinetto,
  Parameter-free online test-time adaptation,
  In \textit{{Proceedings of the IEEE/CVF Conference on Computer Vision and Pattern Recognition (CVPR)}}, pages 8344--8353, 2022.

\bibitem[Senqiao(2024)]{yang2024exploring}
  Senqiao Yang,Jiarui Wu, Jiaming Liu, Xiaoqi Li, Qizhe Zhang, Mingjie Pan, Yulu Gan, Zehui Chen, and Shanghang Zhang,
  Exploring sparse visual prompt for domain adaptive dense prediction,
  In \textit{{Proceedings of International Conference on Artificial Intelligence (AAAI)}},vol. 38, no. 15, pages 16334--16342, 2024.

\bibitem[Jiayi(2024)]{ni2024distribution}
  Jiayi Ni,Senqiao Yang, Ran Xu, Jiaming Liu, Xiaoqi Li, Wenyu Jiao, Zehui Chen, Yi Liu, and Shanghang Zhang,
  Distribution-Aware Continual Test-Time Adaptation for Semantic Segmentation,
  In \textit{{IEEE International Conference on Robotics and Automation (ICRA)}}, pages 3044--3050, 2024.

\bibitem[Lan-Zhe(2022)]{guo2022class}
  Lan-Zhe Guo and Yu-Feng Li,
  Class-imbalanced semi-supervised learning with adaptive thresholding,
  In \textit{{Proceedings of International Conference on Machine Learning (ICML)}}, pages 8082--8094, 2022.

\bibitem[M.(1978)]{smith1978environmental}
  Steven M. Smith,Arthur Glenberg, and Robert A. Bjork,
  Environmental context and human memory,
  \textit{Memory \& Cognition}, vol. 6, no. 4, pages 342--353, Springer, 1978.

\bibitem[Audrey(2017)]{maille2017ecophysiology}
  Audrey Maille and Carsten Schradin,
  Ecophysiology of cognition: How do environmentally induced changes in physiology affect cognitive performance?,
  \textit{Biological Reviews}, vol. 92, no. 2, pages 1101--1112, Wiley Online Library, 2017.

\bibitem[Maxime(2020)]{cauchoix2020cognition}
  Maxime Cauchoix,Alexis S Chaine, and Gladys Barragan-Jason,
  Cognition in context: plasticity in cognitive performance in response to ongoing environmental variables,
  \textit{Frontiers in Ecology and Evolution}, vol. 8, page 106, Frontiers Media SA, 2020.

\bibitem[Qin(2022)]{wang2022continual}
  Qin Wang,Olga Fink, Luc Van Gool, and Dengxin Dai,
  Continual test-time domain adaptation,
  In \textit{{Proceedings of the IEEE/CVF Conference on Computer Vision and Pattern Recognition (CVPR)}}, pages 7201--7211, 2022.

\bibitem[Nikita(2021)]{araslanov2021self}
  Nikita Araslanov and Stefan Roth,
  Self-supervised augmentation consistency for adapting semantic segmentation,
  In \textit{{Proceedings of the IEEE/CVF Conference on Computer Vision and Pattern Recognition (CVPR)}}, pages 15384--15394, 2021.

\bibitem[Ke(2020)]{mei2020instance}
  Ke Mei,Chuang Zhu, Jiaqi Zou, and Shanghang Zhang,
  Instance adaptive self-training for unsupervised domain adaptation,
  In \textit{{Proceedings of the European Conference on Computer Vision (ECCV)}}, pages 415--430, 2020.

\bibitem[Jian(2020)]{liang2020we}
  Jian Liang,Dapeng Hu, and Jiashi Feng,
  Do we really need to access the source data? source hypothesis transfer for unsupervised domain adaptation,
  In \textit{{Proceedings of International Conference on Machine Learning (ICML)}}, pages 6028--6039, 2020.

\bibitem[Dequan(2021)]{wang2020tent}
  Dequan Wang,Evan Shelhamer, Shaoteng Liu, Bruno Olshausen, and Trevor Darrell,
  Tent: Fully test-time adaptation by entropy minimization,
  In \textit{{Proceedings of International Conference on Learning Representations (ICLR) }}, 2021.

\bibitem[Yu(2020)]{sun2020test}
  Yu Sun,Xiaolong Wang, Zhuang Liu, John Miller, Alexei Efros, and Moritz Hardt,
  Test-time training with self-supervision for generalization under distribution shifts,
  In \textit{{Proceedings of International Conference on Machine Learning (ICML)}}, pages 9229--9248, 2020.

\bibitem[Mario(2023)]{dobler2023robust}
  Mario D{\"obler},Robert A Marsden, and Bin Yang,
  Robust mean teacher for continual and gradual test-time adaptation,
  In \textit{{Proceedings of the IEEE/CVF Conference on Computer Vision and Pattern Recognition (CVPR)}}, pages 7704--7714, 2023.

\bibitem[Hojoon(2025)]{lee2025prototypical}
  Hojoon Lee,Seunghwan Lee, Inyoung Jung, and Sungeun Hong,
  Prototypical class-wise test-time adaptation,
  \textit{Pattern Recognition Letters}, vol. 187, pages 49--55, Elsevier, 2025.

\bibitem[Dhanajit(2023)]{brahma2023probabilistic}
  Dhanajit Brahma and Piyush Rai,
  A Probabilistic Framework for Lifelong Test-Time Adaptation,
  In \textit{{Proceedings of the IEEE/CVF Conference on Computer Vision and Pattern Recognition (CVPR)}}, pages 3582--3591, 2023.

\bibitem[Chuan(2017)]{guo2017calibration}
  Chuan Guo,Geoff Pleiss, Yu Sun, and Kilian Q Weinberger,
  On calibration of modern neural networks,
  In \textit{{Proceedings of International Conference on Machine Learning (ICML)}}, pages 1321--1330, 2017.

\bibitem[Chaoqi(2019)]{chen2019progressive}
  Chaoqi Chen,Weiping Xie, Wenbing Huang, Yu Rong, Xinghao Ding, Yue Huang, Tingyang Xu, and Junzhou Huang,
  Progressive feature alignment for unsupervised domain adaptation,
  In \textit{{Proceedings of the IEEE/CVF Conference on Computer Vision and Pattern Recognition (CVPR)}}, pages 627--636, 2019.

\bibitem[Tsung-Yi(2017)]{lin2017focal}
  Tsung-Yi Lin,Priya Goyal, Ross Girshick, Kaiming He, and Piotr Doll{\'a}r,
  Focal loss for dense object detection,
  In \textit{{Proceedings of International Conference on Computer Vision (ICCV)}}, pages 2980--2988, 2017.

\bibitem[Marius(2016)]{cordts2016cityscapes}
  Marius Cordts,Mohamed Omran, Sebastian Ramos, Timo Rehfeld, Markus Enzweiler, Rodrigo Benenson, Uwe Franke, Stefan Roth, and Bernt Schiele,
  The cityscapes dataset for semantic urban scene understanding,
  In \textit{{Proceedings of the IEEE/CVF Conference on Computer Vision and Pattern Recognition (CVPR)}}, pages 3213--3223, 2016.

\bibitem[Christos(2021)]{sakaridis2021acdc}
  Christos Sakaridis,Dengxin Dai, and Luc Van Gool,
  ACDC: The adverse conditions dataset with correspondences for semantic driving scene understanding,
  In \textit{{Proceedings of International Conference on Computer Vision (ICCV)}}, pages 10765--10775, 2021.

\bibitem[Christoph(2020)]{kamann2020benchmarking}
  Christoph Kamann and Carsten Rother,
  Benchmarking the robustness of semantic segmentation models,
  In \textit{{Proceedings of the IEEE/CVF Conference on Computer Vision and Pattern Recognition (CVPR)}}, pages 8828--8838, 2020.

\bibitem[Tao(2022)]{sun2022shift}
  Tao Sun,Mattia Segu, Janis Postels, Yuxuan Wang, Luc Van Gool, Bernt Schiele, Federico Tombari, and Fisher Yu,
  SHIFT: a synthetic driving dataset for continuous multi-task domain adaptation,
  In \textit{{Proceedings of the IEEE/CVF Conference on Computer Vision and Pattern Recognition (CVPR)}}, pages 21371--21382, 2022.

\bibitem[Junha(2023)]{song2023ecotta}
  Junha Song,Jungsoo Lee, In So Kweon, and Sungha Choi,
  EcoTTA: Memory-Efficient Continual Test-time Adaptation via Self-distilled Regularization,
  In \textit{{Proceedings of the IEEE/CVF Conference on Computer Vision and Pattern Recognition (CVPR)}}, pages 11920--11929, 2023.

\bibitem[Liang(2019)]{du2019ssf}
  Liang Du,Jingang Tan, Hongye Yang, Fianfeng Feng, Xiangyang Xue, Qibao Zheng, Xiaoqing Ye, and Xiaolin Zhang,
  Ssf-dan: Separated semantic feature based domain adaptation network for semantic segmentation,
  In \textit{{Proceedings of International Conference on Computer Vision (ICCV)}}, pages 982--991, 2019.

\bibitem[Yang(2018)]{zou2018cbst}
  Yang Zou,Zhiding Yu, B.V.K. Vijaya Kumar, and Jinsong Wang,
  Unsupervised domain adaptation for semantic segmentation via class-balanced self-training,
  In \textit{{Proceedings of the European Conference on Computer Vision (ECCV)}}, pages 289--305, 2018.

\bibitem[Yang(2019)]{zou2019crst}
  Yang Zou,Zhiding Yu, Xiaofeng Liu, B.V.K. Vijaya Kumar, and Jinsong Wang,
  Confidence regularized self-training,
  In \textit{{Proceedings of International Conference on Computer Vision (ICCV)}}, pages 5982--5991, 2019.

\bibitem[Enze(2021)]{xie2021segformer}
  Enze Xie,Wenhai Wang, Zhiding Yu, Anima Anandkumar, Jose M Alvarez, and Ping Luo,
  SegFormer: Simple and efficient design for semantic segmentation with transformers,
  In \textit{{Proceedings of Neural Information Processing Systems (NeurIPS)}}, vol. 34, pages 12077--12090, 2021.

\bibitem[Liang-Chieh(2017)]{chen2017rethinking}
  Liang-Chieh Chen,George Papandreou, Florian Schroff, and Hartwig Adam,
  Rethinking atrous convolution for semantic image segmentation,
  \textit{arXiv preprint arXiv:1706.05587}, 2017.

\bibitem[P.(2014)]{kingma2014adam}
  Diederik P. Kingma and Jimmy Ba,
  Adam: A method for stochastic optimization,
  \textit{arXiv preprint arXiv:1412.6980}, 2014.

\bibitem[Yunhe(2022)]{gao2022visual}
  Yunhe Gao,Xingjian Shi, Yi Zhu, Hao Wang, Zhiqiang Tang, Xiong Zhou, Mu Li, and Dimitris N Metaxas,
  Visual prompt tuning for test-time domain adaptation,
  \textit{arXiv preprint arXiv:2210.04831}, 2022.

\bibitem[Yulu(2023)]{gan2023decorate}
  Yulu Gan,Yan Bai, Yihang Lou, Xianzheng Ma, Renrui Zhang, Nian Shi, and Lin Luo,
  Decorate the newcomers: Visual domain prompt for continual test time adaptation,
  In \textit{{Proceedings of International Conference on Artificial Intelligence (AAAI)}}, vol. 37, no. 6, pages 7595--7603, 2023.

\bibitem[Jiaming(2024)]{liu2024vida}
  Jiaming Liu,Senqiao Yang, Peidong Jia, Ming Lu, Yandong Guo, Wei Xue, and Shanghang Zhang,
  Vi{DA: Homeostatic Visual Domain Adapter for Continual Test Time Adaptation},
  In \textit{{Proceedings of International Conference on Learning Representations (ICLR) }}, 2024.

\bibitem[Kihyuk(2020)]{sohn2020fixmatch}
  Kihyuk Sohn,David Berthelot, Chun-Liang Li, Zizhao Zhang, Nicholas Carlini, Ekin D. Cubuk, Alex Kurakin, Han Zhang, and Colin Raffel,
  FixMatch: Simplifying Semi-Supervised Learning with Consistency and Confidence,
  In \textit{{Proceedings of Neural Information Processing Systems (NeurIPS)}}, vol. 33, pages 596--608, 2020.

\bibitem[Bowen(2021)]{zhang2021flexmatch}
  Bowen Zhang,Yidong Wang, Wenxin Hou, Hao Wu, Jindong Wang, Manabu Okumura, and Takahiro Shinozaki,
  FlexMatch: Boosting Semi-Supervised Learning with Curriculum Pseudo Labeling,
  In \textit{{Proceedings of Neural Information Processing Systems (NeurIPS)}}, vol. 34, pages 18408--18419, 2021.

\bibitem[Yidong(2022)]{wang2022freematch}
  Yidong Wang,Hao Chen, Qiang Heng, Wenxin Hou, Yue Fan, Zhen Wu, Jindong Wang, Marios Savvides, Takahiro Shinozaki, Bhiksha Raj, Bernt Schiele, and Xing Xie,
  FreeMatch: Self-Adaptive Thresholding for Semi-Supervised Learning,
  In \textit{{Proceedings of International Conference on Learning Representations (ICLR) }}, 2022.

\bibitem[Debasmit(2023)]{das2023transadapt}
  Debasmit Das,Shubhankar Borse, Hyojin Park, Kambiz Azarian, Hong Cai, Risheek Garrepalli, and Fatih Porikli,
  Transadapt: A Transformative Framework for Online Test Time Adaptive Semantic Segmentation,
  In \textit{{IEEE International Conference on Acoustics},Speech and Signal Processing (ICASSP)}, 2023.

\bibitem[Antti(2017)]{tarvainen2017mean}
  Antti Tarvainen and Harri Valpola,
  Mean teachers are better role models: Weight-averaged consistency targets improve semi-supervised deep learning results,
  In \textit{{Proceedings of Neural Information Processing Systems (NeurIPS)}}, vol. 30, 2017.

\bibitem[Ziquan(2024)]{wang2024exploring}
  Ziquan Wang,Yongsheng Zhang, Zhenchao Zhang, Zhipeng Jiang, Ying Yu, Lei Li, and Lei Zhang,
  Exploring Uncertainty-Based Self-Prompt for Test-Time Adaptation Semantic Segmentation in Remote Sensing Images,
  \textit{Remote Sensing}, vol. 16, no. 7, page 1239, MDPI, 2024.

\bibitem[Kaiming(2016)]{he2016deep}
  Kaiming He,Xiangyu Zhang, Shaoqing Ren, and Jian Sun,
  Deep residual learning for image recognition,
  In \textit{{Proceedings of the IEEE/CVF Conference on Computer Vision and Pattern Recognition (CVPR)}}, pages 770--778, 2016.

\bibitem[Riccardo(2022)]{volpi2022road}
  Riccardo Volpi,Pau De Jorge, Diane Larlus, and Gabriela Csurka,
  On the road to online adaptation for semantic image segmentation,
  In \textit{{Proceedings of the IEEE/CVF Conference on Computer Vision and Pattern Recognition (CVPR)}}, pages 19184--19195, 2022.

\bibitem[Jiaming(2024)]{liu2024continual}
  Jiaming Liu,Ran Xu, Senqiao Yang, Renrui Zhang, Qizhe Zhang, Zehui Chen, Yandong Guo, and Shanghang Zhang,
  Continual-MAE: Adaptive Distribution Masked Autoencoders for Continual Test-Time Adaptation,
  In \textit{{Proceedings of the IEEE/CVF Conference on Computer Vision and Pattern Recognition (CVPR)}}, pages 28653--28663, 2024.

\bibitem[Jungsoo(2023)]{lee2023towards}
  Jungsoo Lee,Debasmit Das, Jaegul Choo, and Sungha Choi,
  Towards open-set test-time adaptation utilizing the wisdom of crowds in entropy minimization,
  In \textit{{Proceedings of International Conference on Computer Vision (ICCV)}}, pages 16380--16389, 2023.

\bibitem[Junha(2023)]{song2023test}
  Junha Song,Kwanyong Park, InKyu Shin, Sanghyun Woo, Chaoning Zhang, and In So Kweon,
  Test-time Adaptation in the Dynamic World with Compound Domain Knowledge Management,
  In \textit{{IEEE International Conference on Robotics and Automation (ICRA)}}, IEEE, 2023.

\bibitem[Christos(2019)]{sakaridis2019dark}
  Christos Sakaridis,Dengxin Dai, and Luc van Gool,
  Guided Curriculum Model Adaptation and Uncertainty-Aware Evaluation for Semantic Nighttime Image Segmentation,
  In \textit{{Proceedings of International Conference on Computer Vision (ICCV)}}, pages 7374--7383, 2019.

\bibitem[Fisher(2020)]{yu2020bdd}
  Fisher Yu,Haofeng Chen, Xin Wang, Wenqi Xian, Yingying Chen, Fangchen Liu, Vashisht Madhavan, and Trevor Darrell,
  BDD100K: A Diverse Driving Dataset for Heterogeneous Multitask Learning,
  In \textit{{Proceedings of the IEEE/CVF Conference on Computer Vision and Pattern Recognition (CVPR)}}, pages 2636--2645, 2020.

\bibitem[Chaoqi(2019)]{chen2019progressive}
  Chaoqi Chen,Weiping Xie, Wenbing Huang, Yu Rong, Xinghao Ding, Yue Huang, Tingyang Xu, and Junzhou Huang,
  Progressive feature alignment for unsupervised domain adaptation,
  In \textit{{Proceedings of the IEEE/CVF Conference on Computer Vision and Pattern Recognition (CVPR)}}, pages 627--636, 2019.

\bibitem[Yanshuo(2024)]{wang2024continual}
  Yanshuo Wang,Jie Hong, Ali Cheraghian, Shafin Rahman, David Ahmedt-Aristizabal, Lars Petersson, and Mehrtash Harandi,
  Continual test-time domain adaptation via dynamic sample selection,
  In \textit{{Proceedings of Winter Conference on Application of Computer Vision (ECCV)}}, pages 1701--1710, 2024.

\bibitem[Xu(2024)]{yang2024versatile}
  Xu Yang,Xuan Chen, Moqi Li, Kun Wei, and Cheng Deng,
  A Versatile Framework for Continual Test-Time Domain Adaptation: Balancing Discriminability and Generalizability,
  In \textit{{Proceedings of the IEEE/CVF Conference on Computer Vision and Pattern Recognition (CVPR)}}, pages 23731--23740, 2024.

\bibitem[Samarth(2023)]{sinha2023test}
  Samarth Sinha,Peter Gehler, Francesco Locatello, and Bernt Schiele,
  Test: Test-time self-training under distribution shift,
  In \textit{{Proceedings of Winter Conference on Application of Computer Vision (ECCV)}}, pages 2759--2769, 2023.

\bibitem[Yushu(2023)]{li2023robust}
  Yushu Li,Xun Xu, Yongyi Su, and Kui Jia,
  On the Robustness of Open-World Test-Time Training: Self-Training with Dynamic Prototype Expansion,
  In \textit{{Proceedings of International Conference on Computer Vision (ICCV)}}, pages 11836--11846, 2023.

\bibitem[Xu(2023)]{yang2023exlore}
  Xu Yang,Yanan Gu, Kun Wei, and Cheng Deng,
  Exploring Safety Supervision for Continual Test-time Domain Adaptation,
  In \textit{Proceedings of International Joint Conference on Artificial Intelligence (IJCAI)}, pages 1649--1657, 2023.

\bibitem[Jungsoo(2023)]{lee2023crowds}
  Jungsoo Lee,Debasmit Das, Jaegul Choo, and Sungha Choi,
  Towards Open-Set Test-Time Adaptation Utilizing the Wisdom of Crowds in Entropy Minimization,
  In \textit{{Proceedings of International Conference on Computer Vision (ICCV)}}, pages 16380--16389, 2023.

\bibitem[Gilad(2024)]{cohen2024simple}
  Gilad Cohen and Raja Giryes,
  Simple Post-Training Robustness using Test Time Augmentations and Random Forest,
  In \textit{Proceedings of the IEEE/CVF Winter Conference on Applications of Computer Vision (WACV)}, pages 3996--4006, 2024.

\bibitem[Daeun(2024)]{lee2024becotta}
  Daeun Lee,Jaehong Yoon, and Sung Ju Hwang,
  BECoTTA: Input-dependent Online Blending of Experts for Continual Test-time Adaptation,
  \textit{arXiv preprint arXiv:2402.08712}, 2024.

\bibitem[Sergey(2015)]{ioffe2015batch}
  Sergey Ioffe and Christian Szegedy,
  Batch normalization: Accelerating deep network training by reducing internal covariate shift,
  In \textit{{Proceedings of International Conference on Machine Learning (ICML)}}, pages 448--456, 2015.

\bibitem[Marvin(2022)]{zhang2022memo}
  Marvin Zhang,Sergey Levine, and Chelsea Finn,
  Memo: Test time robustness via adaptation and augmentation,
  In \textit{{Proceedings of Neural Information Processing Systems (NeurIPS)}}, vol. 35, pages 38629--38642, 2022.

\bibitem[Sachin(2022)]{goyal2022conjugate}
  Sachin Goyal,Mingjie Sun, Aditi Raghunathan, and Zico Kolter,
  Test-Time Adaptation via Conjugate Pseudo-Labels,
  In \textit{{Proceedings of Neural Information Processing Systems (NeurIPS)}}, vol. 35, pages 6204--6218, 2022.

\bibitem[Yossi(2022)]{gandelsman2022test}
  Yossi Gandelsman,Yu Sun, Xinlei Chen, and Alexei Efros,
  Test-time training with masked autoencoders,
  In \textit{{Proceedings of Neural Information Processing Systems (NeurIPS)}}, vol. 35, pages 29374--29385, 2022.

\bibitem[Longhui(2023)]{yuan2023robust}
  Longhui Yuan,Binhui Xie, and Shuang Li,
  Robust test-time adaptation in dynamic scenarios,
  In \textit{{Proceedings of the IEEE/CVF Conference on Computer Vision and Pattern Recognition (CVPR)}}, pages 15922--15932, 2023.

\bibitem[J(2021)]{hu2021lora}
  Edward J Hu,Yelong Shen, Phillip Wallis, Zeyuan Allen-Zhu, Yuanzhi Li, Shean Wang, Lu Wang, and Weizhu Chen,
  Lora: Low-rank adaptation of large language models,
  \textit{arXiv preprint arXiv:2106.09685}, 2021.

\bibitem[Dan(2019)]{hendrycks2018benchmarking}
  Dan Hendrycks and Thomas Dietterich,
  Benchmarking Neural Network Robustness to Common Corruptions and Perturbations,
  In \textit{{Proceedings of International Conference on Learning Representations (ICLR)}}, 2019.

\bibitem[R.(2016)]{richter2016gta}
  Stephan R. Richter,Vibhav Vineet, Stefan Roth, and Vladlen Koltun,
  Playing for data: Ground truth from computer games,
  In \textit{{Proceedings of the European Conference on Computer Vision (ECCV)}}, 2016.

\bibitem[Yawei(2024)]{luo2024kill}
  Yawei Luo,Ping Liu, and Yi Yang,
  Kill two birds with one stone: Domain generalization for semantic segmentation via network pruning,
  \textit{International Journal of Computer Vision}, pages 1--18, Springer, 2024.

\bibitem[Xin(2024)]{luo2024crots}
  Xin Luo,Wei Chen, Zhengfa Liang, Longqi Yang, Siwei Wang, and Chen Li,
  Crots: Cross-domain teacher--student learning for source-free domain adaptive semantic segmentation,
  \textit{International Journal of Computer Vision,} vol. 132, no. 1, pages 20--39, Springer, 2024.

\bibitem[Zixin(2024)]{wang2024search}
  Zixin Wang,Yadan Luo, Liang Zheng, Zhuoxiao Chen, Sen Wang, and Zi Huang,
  In search of lost online test-time adaptation: A survey,
  \textit{International Journal of Computer Vision}, pages 1--34, Springer, 2024.

\bibitem[Zhilin(2024)]{zhu2024reshaping}
  Zhilin Zhu,Xiaopeng Hong, Zhiheng Ma, Weijun Zhuang, Yaohui Ma, Yong Dai, and Yaowei Wang,
  Reshaping the Online Data Buffering and Organizing Mechanism for Continual Test-Time Adaptation,
  In \textit{{Proceedings of the European Conference on Computer Vision (ECCV)}}, pages 415--433, 2024.

\bibitem[Rongyu(2024)]{zhang2024decomposing}
  Rongyu Zhang,Aosong Cheng, Yulin Luo, Gaole Dai, Huanrui Yang, Jiaming Liu, Ran Xu, Li Du, Yuan Du, Yanbing Jiang, and Shanghang Zhang,
  Decomposing the Neurons: Activation Sparsity via Mixture of Experts for Continual Test Time Adaptation,
  \textit{arXiv preprint arXiv:2405.16486}, 2024.

\end{thebibliography}
\end{document}